%% file: neurips_2026.tex
\documentclass{article}
\PassOptionsToPackage{table}{xcolor}

\usepackage[main, final]{neurips_2026}

\usepackage[utf8]{inputenc} 
\usepackage[T1]{fontenc}    
\usepackage{hyperref}       
\usepackage{url}            
\usepackage{booktabs}       
\usepackage{amsfonts}       
\usepackage{nicefrac}       
\usepackage{microtype}      
\usepackage{xcolor}         
\usepackage{booktabs}
\usepackage{multirow}
\usepackage{graphicx} 

\input{preamble}
\title{HAPS: \underline{H}ierarchical LLM Routing with Joint\\\underline{A}rchitecture and \underline{P}arameter \underline{S}earch}

\author{
  \textbf{Zihang Tian}$^{1}$,
  \textbf{Rui Li}$^{1}$,
  \textbf{Jingsen Zhang}$^{1}$,
  \textbf{Xiaohe Bo}$^{1}$,
  \textbf{Wei Huo}$^{2}$,
  \textbf{Xu Chen}$^{1,*}$\\
  $^{1}$Renmin University of China\\
  $^{2}$Wireless Technology Lab, Huawei Technologies Co., Ltd.\\
  \texttt{\{zihangtian, xu.chen\}@ruc.edu.cn}
}

\begin{document}

\maketitle
\begingroup
  \renewcommand{\thefootnote}{*}
  \footnotetext[1]{Corresponding author.}
\endgroup
\setcounter{footnote}{0}
\input{sections/abstract}

\input{sections/intro}

\input{sections/preliminaries}

\input{sections/method}

\input{sections/exp/setup}

\input{sections/exp/overall}

\input{sections/exp/component}

\input{sections/exp/architecture}

\input{sections/exp/scaling}

\input{sections/exp/tradeoff}

\input{sections/exp/mixed}

\input{sections/related}

\input{sections/limitations}

\input{sections/conclusion}

\input{sections/acknowledgments}
\bibliographystyle{plainnat}
\bibliography{main}

\appendix
\clearpage
\input{sections/appendix/architecture}

\input{sections/appendix/mas_setup}

\input{sections/appendix/data}

\input{sections/appendix/training}

\input{sections/appendix/ifeval}

\input{sections/appendix/efficiency}

\input{sections/appendix/tradeoff}

\input{sections/appendix/case_study}

\input{sections/appendix/prompts}

\clearpage
\input{sections/appendix/broader_impacts}

\clearpage
\input{checklist.tex}

\end{document}

%% file: preamble.tex
\usepackage[utf8]{inputenc}
\usepackage[T1]{fontenc}
\usepackage{microtype}      
\usepackage{times}          
\usepackage{latexsym}
\usepackage{xurl}
\usepackage{amsmath}
\usepackage{amssymb}
\usepackage{amsfonts}
\usepackage{mathtools}      
\usepackage{amsthm}
\usepackage{mathrsfs}       
\usepackage{nicefrac}       

\theoremstyle{definition}

\theoremstyle{remark}
\newtheorem{remark}{\textbf{Remark}}

\usepackage{booktabs}       
\usepackage{multirow}
\usepackage{tabularx}       
\usepackage{makecell}       
\usepackage{graphicx}       
\usepackage{wrapfig}        
\usepackage{needspace}      
\usepackage{adjustbox}      
\usepackage{float}          

\usepackage[ruled,linesnumbered,vlined]{algorithm2e} 

\SetKwComment{tcc}{// }{}   
\SetKwInput{KwIn}{Input}    
\SetKwInput{KwOut}{Output}
\SetKw{KwRet}{return}
\SetKw{KwBreak}{break}
\SetAlgoCaptionSeparator{:} 

\usepackage[inline]{enumitem} 
\usepackage{balance}          

\usepackage{hyperref}         
\usepackage{url}
\usepackage[capitalise]{cleveref} 

\definecolor{headerGray}{gray}{0.95}

\newcolumntype{Y}{>{\centering\arraybackslash}X}
\usepackage[most]{tcolorbox}
\tcbuselibrary{listings, breakable} 

\definecolor{retroBlue}{RGB}{68, 108, 169}   
\newtcblisting{PromptBox}[2][]{
    enhanced,
    breakable,
    listing only,
    listing options={
        basicstyle=\small\rmfamily,
        breaklines=true,
        breakatwhitespace=false,
        columns=fullflexible,
        aboveskip=0pt, belowskip=0pt,
        escapechar=@,
        breakindent=0pt,     
        breakautoindent=false 
    },
    colframe=retroBlue,
    colback=white,
    coltitle=white,
    colbacktitle=retroBlue,
    title={#2},
    fonttitle=\bfseries\large,
    boxrule=1.2pt,
    sharp corners,
    attach boxed title to top left={yshift=-0mm, xshift=0mm},
    boxed title style={sharp corners, boxrule=0pt, bottomrule=1pt},
    #1
}

%% file: sections/abstract.tex
\begin{abstract}
Large language model (LLM) routing aims to exploit the specialized strengths of different LLMs for diverse tasks.
However, existing approaches typically focus only on selecting LLM architectures, treating candidate models as fixed while overlooking parameter adaptation, which can substantially affect task performance.
In this paper, we introduce \textbf{HAPS}, a hierarchical LLM routing framework that \emph{jointly searches over model architectures and parameters}. 
Specifically, a high-level router selects among candidate LLM architectures, while a low-level router generates input-conditioned LoRA parameters for the selected architecture, expanding each candidate from a single fixed behavior into a region of reachable behaviors.
The low-level router is built on the backbone of the high-level router, so that architecture selection and parameter generation share a common representation of the prompt.
We further optimize the whole framework with a reinforcement learning objective.
Extensive experiments show that HAPS consistently outperforms strong routing baselines across reasoning- and knowledge-intensive benchmarks, with analyses of its components, design choices, scalability, cost, and mixed open- and closed-source deployment further demonstrating the robustness of these gains.
Our code is available at \url{https://github.com/zihangtian/HAPS}.
\end{abstract}

%% file: sections/intro.tex
\section{Introduction}
\begin{figure}[t]
\centering
\includegraphics[trim={3pt 23pt 3pt 7pt}, clip, width=\textwidth]{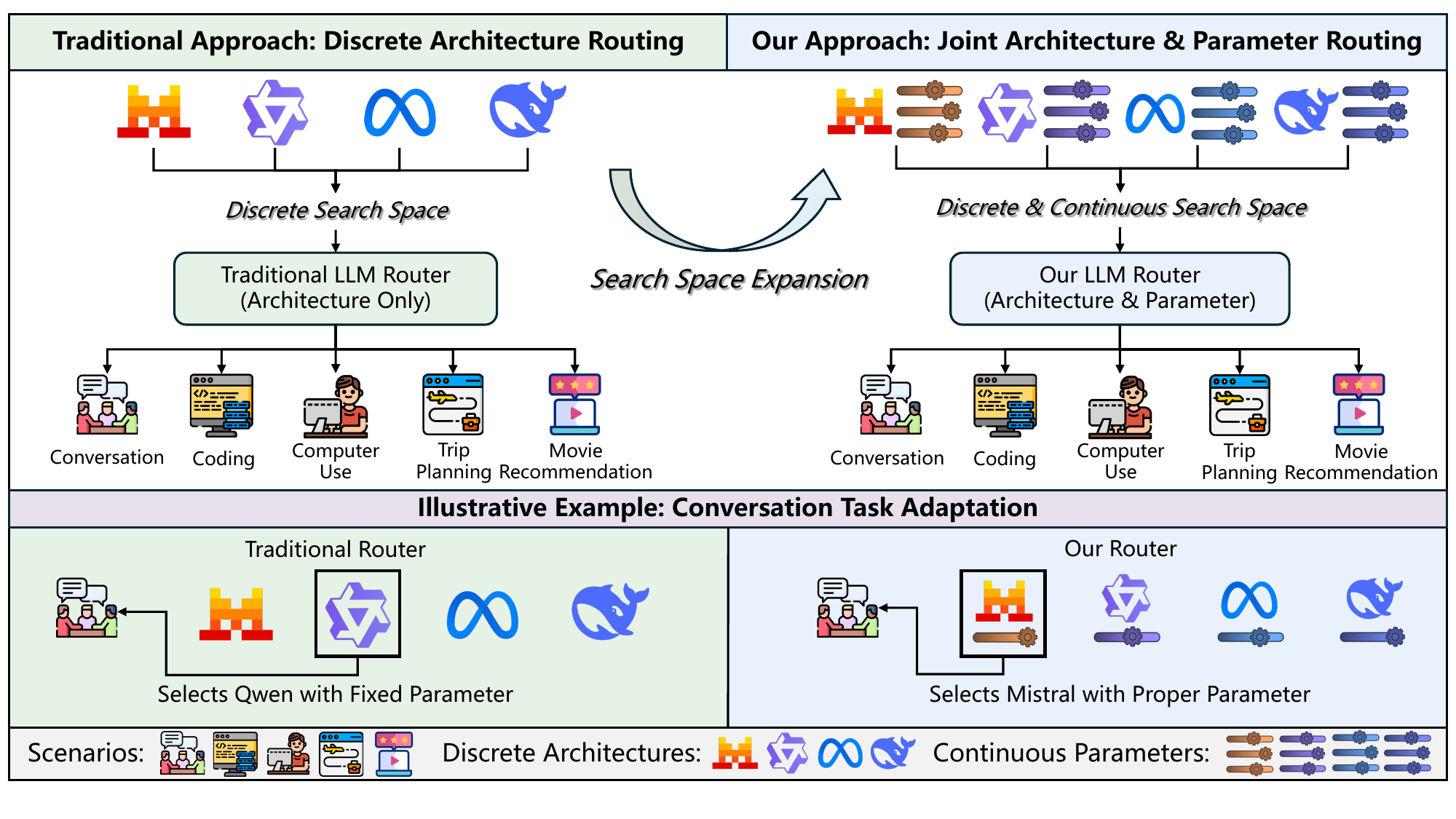}
\vspace{-8pt}
\caption{Illustration of the traditional LLM router versus our proposed model. Unlike standard methods that only select architectures, our framework jointly routes both LLM architectures and their specific parameters, so the preferred architecture may change once its parameters can adapt.}
\label{overall}
\vspace{-12pt}
\end{figure}
Large language models (LLMs) have achieved success across a wide range of natural language processing tasks~\citep{brown2020language, raffel2020exploring}.
However, due to differences in their architectures and parameters, individual LLMs often exhibit distinct strengths and weaknesses. 
To better leverage the complementary capabilities of different models, LLM routing has recently attracted increasing attention, aiming to dynamically select the most appropriate LLM for different tasks.

LLM routing typically starts by defining a pool of candidate architectures (e.g., Llama~\citep{touvron2023llama, touvron2023llama2}, Qwen~\citep{bai2023qwen}, Mistral~\citep{jiang2023mistral7b}). A classifier is then used to map each input query to the most appropriate architecture in the pool, which subsequently generates the response.
The router is usually trained with a supervision signal based on the response correctness. 
For example, RouteLLM~\citep{ong2024routellm} trains a binary router that switches between a stronger and a weaker model using preference data and data augmentation.
GraphRouter~\citep{feng2024graphrouter} constructs a heterogeneous graph over tasks, queries, and LLMs, and formulates routing as inductive edge-attribute prediction.
IRT-Router~\citep{song2025irt} explicitly models query difficulty and model ability, yielding interpretable routing decisions and improved cold-start performance.

While the above models have shown promising performance, they all route within a fixed pool of candidates: each candidate is used with its parameters unchanged, and since parameters determine how a model behaves, each candidate offers only a single, fixed behavior, i.e., one \emph{point} in the space of behaviors an LLM can exhibit.
Meanwhile, we posit that each query is best served by a set of ideal behaviors, and routing is in essence an attempt to reach one of them.
The prevailing paradigm is thus, conceptually, nearest-point retrieval over a handful of fixed points: even a perfect router can only pick the point closest to this set, leaving a gap that no classifier, however well trained, can close.

This raises the question from which our work starts: why should a candidate be a single point? Since behavior comes from parameters, the natural objects to route over are the parameters themselves. Pretrained models are already strong, so we keep them as anchors rather than search parameters from scratch, and expand each anchor into a \emph{region} around it by generating an input-conditioned low-rank update on top of the selected architecture. The pool stays the same size while the behavior it covers grows: each candidate now offers a continuum of input-specific configurations rather than a single one. Architectures differ in how far and in which directions such an update can steer them, so the anchor closest to a query by default need not be the one that can reach closest to it. As illustrated in Figure~\ref{overall}, for a conversation task an architecture-only router may settle on Qwen because it is the best of the fixed candidates, whereas Mistral, weaker in its default configuration, turns out to have the greater potential once its parameters are allowed to adapt. Reaching that better solution requires the two decisions to be made together, so we route over architectures and parameters \emph{jointly}.

Although the above idea is promising, its implementation presents significant challenges. Specifically, existing LLM routers typically employ classifiers to select architectures from a discrete space, whereas LLM parameters reside in a continuous space, making them difficult to directly integrate into these routing models. Moreover, since both architecture and parameter search aim to improve the quality of responses for a given query, they should ideally be guided by shared underlying principles. However, existing methods for architecture routing (e.g., RouteLLM) and parameter tuning (e.g., LoRA~\citep{hu2022lora}) follow fundamentally different strategies. Enabling these components to share common knowledge therefore requires deliberate algorithmic and architectural design.

To address the challenges outlined above, we propose a hierarchical LLM router named \textbf{HAPS}. At the high level, a Llama-based classifier is used to select the most suitable LLM architecture for each component of the system, such as each agent in a multi-agent pipeline. At the low level, instead of searching for free-form LLM parameters directly, we generate them through a neural network. With this design, both architecture selection and parameter search can be formulated as prediction problems, providing a common basis for combining them within a unified framework for LLM routing. To enable shared knowledge between architecture selection and parameter search, the low-level router reuses part of the high-level router's parameters and extends them. In the training process, we jointly optimize both high- and low-level routers using reinforcement learning, so that both levels are driven by the same task reward. We conduct extensive experiments to demonstrate the contributions of model components and to evaluate the effectiveness of our model compared to strong routing baselines.
The main contributions of this paper are summarized as follows:
\begin{itemize}[leftmargin=*]
\item We introduce the idea of joint model architecture and parameter search for LLM routing, which lifts each candidate from a fixed point to a region of reachable behaviors.
\item We implement the above idea via a hierarchical framework, designing a parameter generation network that facilitates knowledge sharing between the two levels and leveraging a reinforcement learning objective to enable joint discrete and continuous optimization.
\item We conduct extensive experiments across reasoning and knowledge benchmarks, where HAPS outperforms strong routing baselines, together with analyses that validate its key design choices.
\end{itemize}

%% file: sections/preliminaries.tex
\section{Preliminaries}
\label{pre}
To make our presentation clearer, we first introduce the basic concepts of LLM routing. We define a pool of candidate LLMs as \(\mathcal{M} = \{m_0, \dots, m_{K-1}\}\), where each \(m_k\) denotes an LLM with a distinct architecture (e.g., Llama or Qwen), and \(K\) is the total number of models. Given a system with \(N\) components that require routing (e.g., the agents in a multi-agent pipeline, each handling one subtask), the objective is to learn a classifier \(\mathbf{r}:\mathcal{P} \rightarrow \{0, \dots, K-1\}^{N}\) that maps a routing prompt to the indices of the most suitable LLMs in \(\mathcal{M}\) for generating accurate responses.

Let \(\theta\) denote the parameters of the router \(\mathbf{r}\). The training objective is to minimize the prediction loss between the output generated by the selected models and the ground-truth label \(y\):
\begin{equation}
\label{pre-loss}
\min_{\theta} \;\mathop{\mathbb{E}}\limits_{(p,\mathbf{x},y)\sim\mathcal{D}} \left[ L\left(\mathrm{G}\left(\mathbf{m}_{\mathbf{r}({p};\theta)}, \mathbf{x}\right), y\right) \right],
\end{equation}
where \(\mathcal{D}\) is the data distribution, \({p} \in \mathcal{P}\) is the routing prompt (e.g., the task description, the role of each component, and the list of candidate LLMs), and \(\mathbf{x} {=} \{x_i\}_{i=0}^{N-1} \) collects the inputs issued to the LLMs, one per component. We let \( \mathbf{m}_{\mathbf{r}({p};\theta)}{=}\{ {m}_{[\mathbf{r}({p};\theta)]_i}  \}_{i=0}^{N-1}    \) denote the set of LLMs selected by the router \(\mathbf{r}\). The function \(\mathrm{G}(\mathbf{m}_{\mathbf{r}({p};\theta)}, \mathbf{x})\) represents the system output obtained by applying each input $x_i\in \mathbf{x}$ to its corresponding LLM ${m}_{[\mathbf{r}({p};\theta)]_i}\in \mathbf{m}_{\mathbf{r}({p};\theta)}$. 
Here, \(L(\cdot, \cdot)\) is a task-specific loss.

Equation~(\ref{pre-loss}) has been instantiated in various forms by prior work~\citep{ong2024routellm, feng2024graphrouter, song2025irt}. However, most existing LLM routing methods focus solely on selecting LLM architectures, neglecting the role of parameter adaptation, an important factor for task performance (as illustrated in Figure~\ref{overall}). To address this limitation, in the following section, we introduce a hierarchical LLM router that treats the parameters of each selected LLM as searchable variables and jointly routes the architectures and parameters to enhance task performance.

%% file: sections/method.tex
\section{The HAPS Model}
\label{sec:haps}
\begin{figure*}[t]
\centering
\includegraphics[trim={99pt 189pt 110pt 104pt}, clip, width=\textwidth]{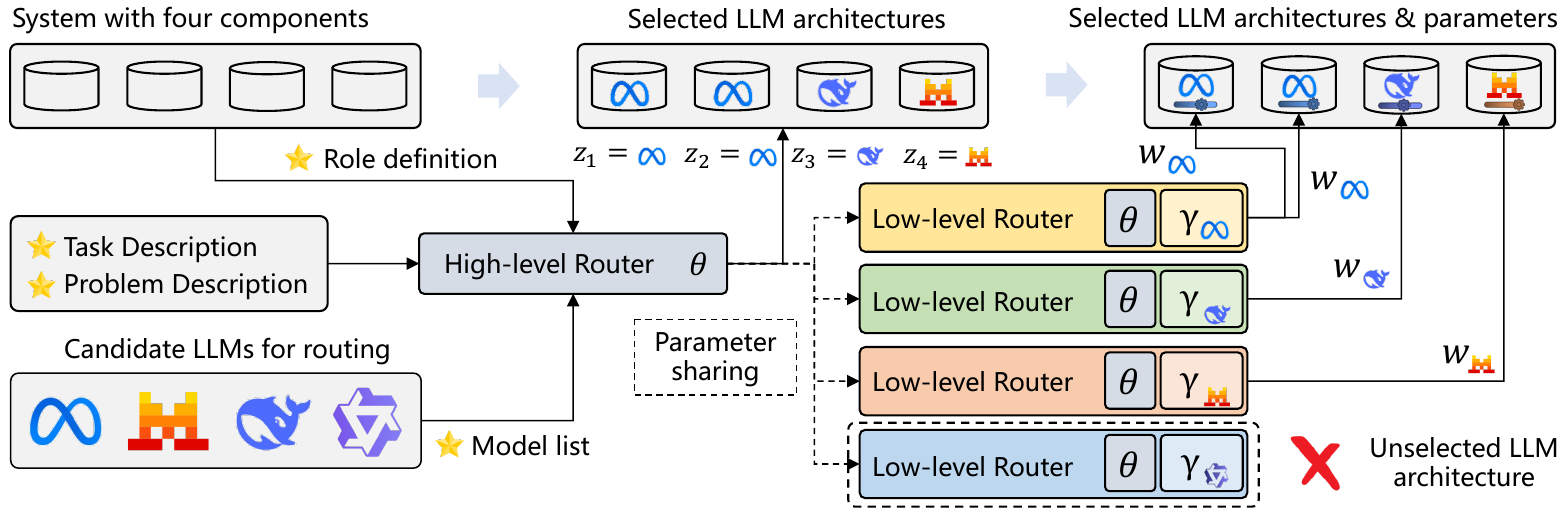}
\vspace{-10pt}
\caption{The overall framework of our hierarchical LLM router.
The complete routing process includes two phases:
(1) Architecture search based on the high-level router.
(2) Parameter search based on the low-level router.
The two routers share the backbone $\theta$ and thus a common prompt representation, and only the head $\gamma_k$ of the selected architecture generates its parameters.
}
\label{model}
\vspace{-5pt}
\end{figure*}

As shown in Figure~\ref{model}, our model adopts a hierarchical structure: a high-level router selects an LLM architecture, and a low-level router generates parameters for the selected architecture as a low-rank update to its frozen weights. Both levels read the same routing prompt, and they share part of their parameters so that knowledge learned for one transfers to the other. We describe each of them below.

\subsection{High-Level Router}
The high-level router is responsible for selecting the appropriate LLM architectures. Intuitively, this selection should be guided by the tasks and roles we expect the LLMs to perform. Accordingly, we define $p$ in Equation~(\ref{pre-loss}) as a four-part routing prompt, structured as follows:
(1) \textit{Role definition} -- a brief textual specification of each agent's capabilities and expected behaviors;
(2) \textit{Task description} -- a concise summary of the overall objective;
(3) \textit{Problem description} -- a detailed description of the specific problem to be addressed by the LLMs (i.e., $\mathbf{x}$ in Equation~(\ref{pre-loss}));
(4) \textit{Model list} -- the names or symbolic identifiers of all candidate LLMs (i.e., $\mathcal{M}$).
Formally, the high-level router $\mathbf{r}$ is a multi-output classifier implemented using a Llama-based architecture, defined as $\mathbf{z} = \mathbf{r}(p;\theta)$, where $\mathbf{z}=\{z_i\}_{i=0}^{N-1}$ is an $N$-dimensional vector, with each element indicating the index of the LLM in $\mathcal{M}$ assigned to the corresponding system component (i.e., $z_i$ means that the $i$th component in the system uses $m_{z_i}$ from $\mathcal{M}$). Here, $\theta$ is the learnable parameter vector of the high-level router. All $N$ entries of $\mathbf{z}$ are produced from one pass over $p$, so the router can assign complementary models to different components, rather than choosing the model for each component independently.

\subsection{Low-Level Router}
\label{sec:low_level}
The low-level router is responsible for finding suitable parameters for the selected LLMs. A straightforward approach assigns free parameters to each prompt and treats parameter search as a supervised optimization task. However, this method has two main limitations. First, assigning free parameters to each prompt implies that the parameters are independent across prompts, making it difficult to generalize the learned parameters to unseen prompts (\textit{low generalization ability}). Second, both architecture search and parameter search aim to enhance task performance, which means they should ideally share underlying knowledge. Using free parameters, however, makes it difficult to establish such a knowledge-sharing mechanism (\textit{poor knowledge-sharing ability}).

To address the above limitations, we propose to construct a neural network $\mathbf{s}$ to generate the LLM parameters.
The parameters of $\mathbf{s}$ serve to compress and relate different prompts within a latent space, thereby allowing better generalization to unseen prompts (\textit{enhancing the generalization ability}).
Moreover, by introducing $\mathbf{s}$, both the architecture and parameter search are unified under a common prediction framework. This unification makes it possible to transfer knowledge through parameter sharing between the two tasks (\textit{enhancing the knowledge-sharing ability}).

Formally, given $K$ candidate LLMs in $\mathcal{M}$, we construct a dedicated parameter generation network $\mathbf{s}_k$ for each LLM $m_k$, since candidate architectures differ in hidden size and therefore require parameters of different shapes. All $\mathbf{s}_k$ take the same form but keep their own layers on top of a shared backbone. To facilitate knowledge sharing, each $\mathbf{s}_k$ is built by stacking several MLP layers on the pooled representation produced by the high-level router $\mathbf{r}$. To keep the generation cost low, $\mathbf{s}_k$ outputs a low-rank update rather than a full weight matrix, giving $w_k = \mathbf{s}_k(p; \phi_k)$, where $p$ is the same routing prompt given to the high-level router, and $w_k$ denotes the generated LoRA parameters tailored for $m_k$. The parameter set $\phi_k {=} \{\theta,\gamma_k\}$ includes $\theta$, the parameters of the high-level router, which are shared across all $\mathbf{s}_k$, and $\gamma_k$, the parameters specific to $\mathbf{s}_k$ and hence to candidate $m_k$.

Sharing $\theta$ serves both levels. The high-level router must already encode which aspects of a prompt separate one candidate from another, since its own decision rests on that, and the same distinction tells the low-level router how the selected candidate should be adjusted. Reusing $\theta$ therefore gives each $\mathbf{s}_k$ a representation informative for parameter generation without learning it again, and keeps the added cost per candidate limited to $\gamma_k$. The two levels also move together during training: gradients from the generated parameters reach $\theta$, so the representation driving architecture selection reflects how well that architecture can be adapted. Section~\ref{sec:component} isolates this effect with a decoupled variant.

\subsection{Model Optimization}
\label{MO}
The routing process of our framework consists of two phases:
(i) Architecture selection, where the high-level router assigns an LLM to each system component, giving the indices $\mathbf{z} = \mathbf{r}(p;\theta)$.
(ii) Parameter search, where for the $i$th component we generate the LoRA parameters of its assigned LLM $m_{z_i}$ as ${w}_{z_i} = \mathbf{s}_{z_i}(p; \phi_{z_i})$.
Finally, we enhance the traditional LLM routing objective in Equation~(\ref{pre-loss}) by allowing the LLM parameters to be searchable, as follows:
\begin{equation}
\label{our-loss}
\min_{\theta, \mathbf{\Gamma}} \;\mathop{\mathbb{E}}\limits_{(p,\mathbf{x},y)\sim\mathcal{D}} \left[ L\left(\mathrm{G}\left(\mathbf{m}_{\mathbf{z}}( \mathbf{w}_{\mathbf{z}} ), \mathbf{x}\right), y\right) \right],
\end{equation}
where $\mathbf{w}_{\mathbf{z}} {=} \{ {w}_{{z}_i} {=} \mathbf{s}_{z_i}(p; \phi_{z_i}) \}_{i=0}^{N-1}$ is the set of the {generated} LoRA parameters for \( \mathbf{m}_{\mathbf{z}}{=} \{ {m}_{{z}_i}  \}_{i=0}^{N-1} \).
The notation \(\mathbf{m}_{\mathbf{z}}(\mathbf{w}_{\mathbf{z}})\) indicates that \({w}_{z_i}\) is added to the {frozen} weights of \({m}_{z_i}\).
Here, $\mathbf{\Gamma}{=}\{\gamma_{k}\}_{k=0}^{K-1}$ collects the generator parameters of all {candidate} LLMs in $\mathcal{M}$, excluding the shared $\theta$.

Equation~(\ref{our-loss}) cannot be optimized directly, since $\mathbf{z}$ is discrete and the task loss is only observed after the selected LLMs have produced their outputs. We therefore optimize it in two stages: we first warm up \(\theta\) on routing decisions that are known to be good, and then jointly optimize \(\theta\) and \(\mathbf{\Gamma}\) with a policy-gradient objective that weights each sampled decision by the reward it receives.

In the warm-up stage, for each prompt \(p\), we sample \( J \) LLM assignment vectors \( \{\mathbf{z}^{(j)}\}_{j=0}^{J-1} \), or enumerate all of them when the number of components and candidates is small, and use each \( \mathbf{z}^{(j)} \) to accomplish the task described in \( p \), where the parameters of all LLMs are kept fixed. We then compute a reward \( R_j \) for each \( \mathbf{z}^{(j)} \) by comparing the final predicted task output with the ground truth, and keep the best-scoring decision as the training sample for initializing \( \theta \), that is, $\{(p,\mathbf{z}^{(j^\star)})\mid j^\star = \arg\max_{j} R_j \}$.
After the warm-up, we replace the task loss with a reward-weighted negative log-likelihood over the routing decisions, which admits REINFORCE-style gradients:
\begin{equation}
\min_{\theta, \mathbf{\Gamma}}
\mathop{\mathbb{E}}\limits_{\substack{(p,y)\sim \mathcal{D}^{\mathrm{RL}}_{\mathrm{train}}}}
\!\left[ -R(\hat{y}, y)\, \log P(\mathbf{z}, \mathbf{w}_{\mathbf{z}} \mid p) \right],
\label{eq:rl-loss-agent}
\end{equation}
where $\mathcal{D}^{\mathrm{RL}}_{\mathrm{train}}$ is the dataset for joint reinforcement learning, drawn as a subset of the warm-up training prompts. $\hat{y} {=} \mathrm{G}(\mathbf{m}_{\mathbf{z}}(\mathbf{w}_{\mathbf{z}}), \mathbf{x})$ is the system output produced under the sampled decision, and $R(\hat{y}, y)$ is a reward function that can take different forms (e.g., task performance or the trade-off between performance and cost). $P(\mathbf{z}, \mathbf{w}_{\mathbf{z}} {\mid} p)$ is the joint probability of the LLM architectures and parameters given the prompt.
The objective thus promotes the decisions that earn higher rewards and suppresses the rest. Both levels are updated from this single reward: $\mathbf{\Gamma}$ directly through the generated parameters, and $\theta$ through the shared backbone on which all $K$ generators are built.

\begin{remark}[\textbf{Extension to mixed open- and closed-source LLM routing}]\label{remark:mix}
While our framework is designed for parameter search, it can also be easily extended to settings involving both open- and closed-source LLMs. A simple 0-1 indicator can be introduced and multiplied with $\mathbf{w}_{\mathbf{z}}$, so that no parameters are injected into closed-source LLMs behind an inference API and no gradients flow to their generators.
Formally, we can refine the objective in Equation~(\ref{our-loss}) as follows:
\begin{equation}
\label{our-loss-imp}
\min_{\theta, \mathbf{\Gamma}} \;\mathop{\mathbb{E}}\limits_{(p,\mathbf{x},y)\sim\mathcal{D}} \left[ L\left(\mathrm{G}\left(\mathbf{m}_{\mathbf{z}}(\mathbf{o}_{\mathbf{z}} \odot \mathbf{w}_{\mathbf{z}} ), \mathbf{x}\right), y\right) \right],
\end{equation}
where $\odot$ is the element-wise product.
$\mathbf{o}_{\mathbf{z}} = \{{o}_{{z}_i}\}_{i=0}^{N-1}$ is an $N$-dimensional 0-1 vector, and ${o}_{{z}_i} = 1$ means that ${m}_{{z}_i}$ is an open-source LLM, so parameter search is confined to those components.
\end{remark}

\begin{remark}[\textbf{Cost of growing the candidate pool}]\label{remark:scale}
Dedicating a generator $\mathbf{s}_k$ to every candidate makes the parameter count grow linearly in $K$, but the slope is small: the backbone is shared, so each candidate only replicates the head on top of it, under $2\%$ of the backbone.
Inference stays batched as well. Since the generated updates are low-rank, they need not be merged into the base weights: the adapter term is evaluated as a separate batched product and added to the shared base computation, so queries with different generated parameters can still share one batched forward pass.
Section~\ref{sec:scaling} reports accuracy and overhead as the pool grows, with implementation details in Appendix~\ref{app:efficiency}.
\end{remark}

%% file: sections/exp/setup.tex
\section{Experiments}
We empirically evaluate our framework around the following research questions (\textbf{RQs}): 
\begin{itemize}[leftmargin=*, topsep=0pt, parsep=0pt, itemsep=0pt]
\item \textbf{RQ1}: Does HAPS achieve superior performance compared to existing LLM routing methods? 
\item \textbf{RQ2}: Are the key components of HAPS necessary for its gains? 
\item \textbf{RQ3}: How do router backbone and LoRA design choices affect HAPS?
\item \textbf{RQ4}: As the candidate pool grows, how do performance and computational overhead change?
\item \textbf{RQ5}: Can HAPS effectively balance the trade-off between task performance and cost?
\item \textbf{RQ6}: Is our framework effective in mixed scenarios involving open- and closed-source LLMs? 
\end{itemize}
In the following, we answer and analyze the above questions in detail based on empirical results.

\subsection{Experimental Setup}

\paragraph{Datasets.}
We evaluate HAPS on two representative benchmarks: \textbf{HotpotQA}~\citep{yang2018hotpotqa} and \textbf{MMLU}~\citep{hendrycks2020measuring}.\footnote{Appendix~\ref{app:ifeval} additionally evaluates HAPS on \textbf{IFEval}~\citep{zhou2023instruction} in a single-agent setting.}
HotpotQA serves as a testbed for multi-hop reasoning, requiring models to synthesize information from multiple supporting facts to answer complex questions.
In contrast, MMLU evaluates broad knowledge coverage across 57 subjects, ranging from STEM to the humanities.
These two datasets allow us to evaluate HAPS from complementary perspectives: complex reasoning generation and broad knowledge selection.
Each benchmark is split into training, validation, and test sets. The warm-up stage trains on the full training set and uses the validation set for checkpoint selection, while the reinforcement learning stage draws a small subset of each. The test set is held out and used only for the final results. Appendix~\ref{app:data_collection} details the splits.

\paragraph{Baselines.}
We compare HAPS against a comprehensive set of routing strategies with complementary inductive biases: (1) \textbf{Random} samples LLMs uniformly; (2) \textbf{RouteLLM}~\citep{ong2024routellm} learns a binary routing policy from preference data and augmentation; (3) \textbf{GraphRouter}~\citep{feng2024graphrouter} models task--query--LLM interactions using a heterogeneous graph; and (4) \textbf{IRT-Router}~\citep{song2025irt} uses item response theory to estimate query difficulty and model ability. To fit our two-agent setting, we apply the protocol-level adjustments summarized in Appendix~\ref{app:data_collection}.

\paragraph{Implementation Details.}
We establish a two-agent environment (i.e., $N{=}2$), which comprises a \textbf{Student} for solution generation and a \textbf{Teacher} for feedback~\citep{bo2024reflective, chen2023agentverse, chen2025optima}.
We evaluate our method across three {open-source} candidate pairs: \textbf{L-Q} (Llama-3.1-8B-Instruct \& Qwen2.5-7B-Instruct), \textbf{M-Q} (Mistral-7B-Instruct-v0.3 \& Qwen2.5-7B-Instruct), and \textbf{L-M} (Llama-3.1-8B-Instruct \& Mistral-7B-Instruct-v0.3). We further extend the evaluation to two {mixed-source} settings in Section~\ref{sec:mixed}: \textbf{L-G} (Llama-3.1-8B-Instruct \& GPT-4.1~Nano) and \textbf{Q-D} (Qwen2.5-7B-Instruct \& DeepSeek~V3~\citep{liu2024deepseek}).
Our high-level router $\mathbf{r}$ employs a Llama-3.2-1B-Instruct backbone to keep routing overhead low.
For each low-level router $\mathbf{s}_k$, we append a parameter generation network, structured as a 2-layer MLP with 256 hidden units, to the backbone.
During training, we empirically set the LoRA scaling factor $\alpha_{\mathrm{lora}} {=} 0.01$ and LoRA rank $r {=} 8$.
The generated LoRA parameters are injected into the output projection (\texttt{o\_proj}) of the final Transformer layer\footnote{RQ3 (Section~\ref{sec:analysis}) analyzes the effects of backbone choice and alternative LoRA injection designs.}.
Following common practice, we use token-level F1 for HotpotQA and accuracy for MMLU.

\begin{table*}[t]
\centering
\small
\renewcommand{\arraystretch}{1.1}
\setlength{\aboverulesep}{0pt}
\setlength{\belowrulesep}{0pt}

\caption{
\textbf{Main Results.}
HAPS compared with routing baselines on HotpotQA and MMLU.
Best and second-best results are bold and underlined.
All values are percentages with ``\%'' omitted.
}
\label{tab:main_results}
\begin{tabularx}{\textwidth}{l YYYYYY}
\toprule
\multirow{2}{*}{\textbf{Method}} &
\multicolumn{3}{c}{\textbf{HotpotQA (F1)}} &
\multicolumn{3}{c}{\textbf{MMLU (Acc)}} \\
\cmidrule(lr){2-4} \cmidrule(lr){5-7}
& \textbf{L-Q} & \textbf{M-Q} & \textbf{L-M}
& \textbf{L-Q} & \textbf{M-Q} & \textbf{L-M} \\
\midrule
Random &
38.94 & 11.44 & 23.26 &
64.48 & 60.00 & 58.97 \\
RouteLLM &
40.67 & \underline{36.10} & \underline{38.95} &
66.55 & \underline{65.86} & 62.76 \\
GraphRouter &
\underline{41.31} & 23.07 & 26.75 &
\underline{66.90} & 65.52 & 63.10 \\
IRT-Router &
37.34 & 34.09 & 38.92 &
65.52 & 65.17 & \underline{63.79} \\
\midrule
{HAPS} &
\textbf{43.16} &
\textbf{39.70} &
\textbf{40.58} &
\textbf{68.97} &
\textbf{67.24} &
\textbf{65.86} \\
\bottomrule
\end{tabularx}
\end{table*}

\begin{table*}[t]
\centering
\small
\renewcommand{\arraystretch}{1.1}
\setlength{\aboverulesep}{0pt}
\setlength{\belowrulesep}{0pt}
\caption{
  \textbf{Ablation Studies.}
  Comparison between HAPS variants and the full model. ``Left'' and ``Right'' denote using the first or second model of the candidate pair (e.g., L in L-Q) as the static assignment applied to every test query.
  Marking and units follow Table~\ref{tab:main_results}.
}
\label{tab:ablation_studies}
\begin{tabularx}{\textwidth}{l YYYYYY}
\toprule
\multirow{2}{*}{\textbf{Variant}} &
\multicolumn{3}{c}{\textbf{HotpotQA (F1)}} &
\multicolumn{3}{c}{\textbf{MMLU (Acc)}} \\
\cmidrule(lr){2-4} \cmidrule(lr){5-7}
& \textbf{L-Q} & \textbf{M-Q} & \textbf{L-M}
& \textbf{L-Q} & \textbf{M-Q} & \textbf{L-M} \\

\midrule
\rowcolor{headerGray}
\multicolumn{7}{l}{\textit{\textbf{High-Level Router Ablation}}} \\

w/o High-Level (Random) &
40.13 & 26.37 & 25.44 &
64.83 & 61.03 & 58.62 \\

w/o High-Level (Fixed): & & & & & & \\

\hspace{3mm} \makebox[2cm][l]{Teacher=Left,} Student=Left &
38.48 & 11.55 & \underline{37.84} &
62.07 & 56.21 & \underline{64.48} \\

\hspace{3mm} \makebox[2cm][l]{Teacher=Right,} Student=Right &
34.23 & \underline{35.51} & 7.10 &
\underline{68.28} & \underline{66.90} & 54.48 \\

\hspace{3mm} \makebox[2cm][l]{Teacher=Left,} Student=Right &
34.06 & 32.25 & 8.89 &
66.90 & 66.55 & 56.55 \\

\hspace{3mm} \makebox[2cm][l]{Teacher=Right,} Student=Left &
\underline{42.45} & 9.54 & 35.04 &
65.52 & 60.69 & 62.76 \\

\midrule
\rowcolor{headerGray}
\multicolumn{7}{l}{\textit{\textbf{Low-Level Router Ablation}}} \\

w/o Low-Level Router &
33.93 & 33.67 & 36.32 &
66.55 & 64.83 & 61.03 \\

\midrule

{HAPS} &
\textbf{43.16} &
\textbf{39.70} &
\textbf{40.58} &
\textbf{68.97} &
\textbf{67.24} &
\textbf{65.86} \\

\bottomrule
\end{tabularx}
\end{table*}

%% file: sections/exp/overall.tex
\subsection{Overall Performance (RQ1)}
We compare \textbf{HAPS} with four baselines in Table~\ref{tab:main_results}. Random is generally weak, while RouteLLM, GraphRouter, and IRT-Router fluctuate across candidate pairs: none is consistently the runner-up, and GraphRouter, second on HotpotQA L-Q, drops to 23.07 F1 on M-Q. This instability suggests that fixed routing criteria do not generalize across pairs; useful routing signals depend on both the models and the input distribution.
HAPS achieves the best performance in all six settings, suggesting that joint search is robust across both complex reasoning generation and broad knowledge selection. On HotpotQA, HAPS outperforms the runner-up by 1.85, 3.60, and 1.63 F1 points on L-Q, M-Q, and L-M, respectively. The largest gain appears on M-Q, the pair on which every baseline scores lowest, suggesting that HAPS helps most when one candidate is much weaker than the other. Here, the margin corresponds to a relative improvement of nearly 10\% over the runner-up, RouteLLM. On MMLU, HAPS improves over the strongest baseline by 2.07, 1.38, and 2.07 accuracy points, respectively. Overall, these gains suggest that jointly searching over architectures and parameters is more flexible than selecting among fixed candidate models without adapting their parameters.

%% file: sections/exp/component.tex
\subsection{Component Analysis (RQ2)}
\label{sec:component}
\begin{wrapfigure}{r}{0.50\columnwidth}
\vspace{-0.8\baselineskip}
  \centering
  \includegraphics[width=0.50\columnwidth]{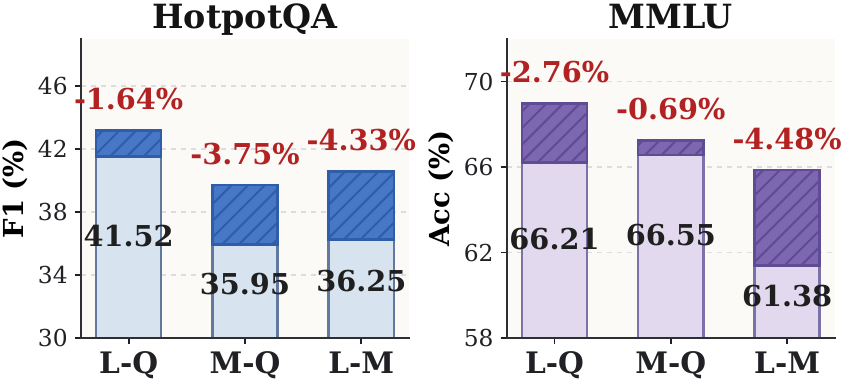}
\vspace{-3pt}
\caption{\textbf{Parameter sharing analysis.} We compare HAPS with its no-sharing variant on HotpotQA and MMLU; red labels mark the performance drop of the no-sharing variant.}
  \label{fig:setting_analysis}
\vspace{-0.8\baselineskip}
\end{wrapfigure}
To identify the sources of HAPS's gains, we compare it with four variants in Table~\ref{tab:ablation_studies} and Figure~\ref{fig:setting_analysis}: (1) \textbf{w/o High-Level (Random)} uses uniform sampling; (2) \textbf{w/o High-Level (Fixed)} enforces static assignments; (3) \textbf{w/o Low-Level} removes LoRA parameter generation; and (4) \textbf{w/o Parameter Sharing} freezes the high-level router after initialization and trains the low-level router with its own backbone. Random selection lowers performance substantially, confirming that a learned high-level policy is necessary. Yet under the same random routing, this variant still outperforms the parameter-free Random baseline in Table~\ref{tab:main_results} in five of six settings (e.g., from 11.44 to 26.37 F1 on HotpotQA M-Q), showing that parameter generation partly compensates for poor routing decisions. Static assignments are also insufficient: even the best one for each pair, chosen in hindsight, falls below HAPS in all six settings, indicating that no single assignment suits every query. Removing the low-level module consistently degrades performance, by up to 9.23 F1 points on HotpotQA, indicating that architecture routing alone cannot fully exploit candidate-specific strengths. Removing parameter sharing also weakens HAPS on every pair, by up to 4.33 F1 points on HotpotQA and 4.48 accuracy points on MMLU, confirming that the two routers benefit from a shared backbone (Section~\ref{sec:low_level}). Appendix~\ref{app:case_study} further toggles only the generated parameters on a fixed routing decision to show how they change the behavior of the selected agents.

%% file: sections/exp/architecture.tex
\subsection{Effects of Architectural Choices (RQ3)}
\label{sec:analysis}
\begin{figure}[t]
\centering
\includegraphics[width=\textwidth]{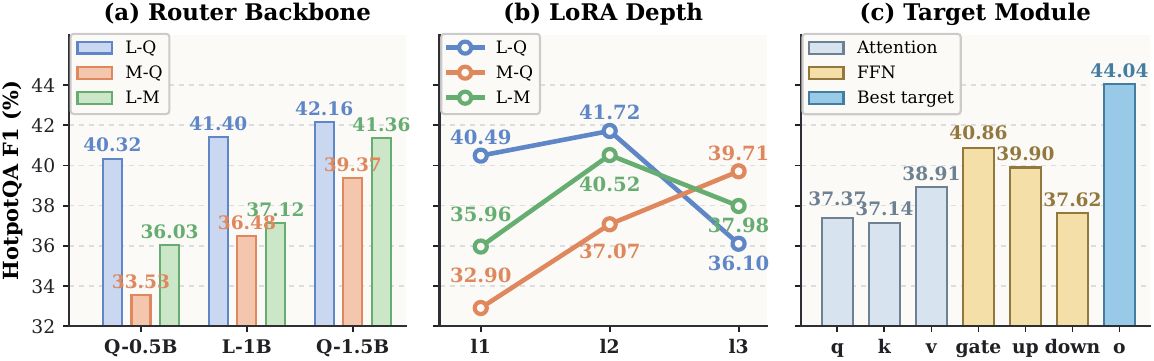}
\caption{
\textbf{Architectural choices in HAPS.}
We test three design factors on HotpotQA: \textbf{(a)} router backbone size, where Q and L denote Qwen and Llama backbones; \textbf{(b)} the number of adapted final layers; and \textbf{(c)} the final-layer target module. Each panel is an independent sweep that varies only the factor shown, with all other hyperparameters held fixed within that panel.
}
\label{fig:analysis}
\vspace{-3.5pt}
\end{figure}

We analyze three architectural choices on HotpotQA: the router backbone, the number of adapted LoRA layers, and the target module in the final layer. Figure~\ref{fig:analysis} shows that router capacity has a clear effect. Scaling the backbone from Qwen2.5-0.5B-Instruct to Llama-3.2-1B-Instruct (our default) and Qwen2.5-1.5B-Instruct raises the average F1 from 36.63 to 38.33 and 40.96, indicating that HAPS benefits from a stronger representation for prompt-level routing. The preferred adaptation depth depends on the candidate pair: L-Q and L-M peak at $\mathbf{l}_2$, while M-Q continues to improve at $\mathbf{l}_3$. Increasing the depth from $\mathbf{l}_1$ to $\mathbf{l}_2$ improves F1 for all three pairs, with gains of up to 4.56 points on L-M. The target-module results further show that \texttt{o\_proj} is the strongest final-layer target on L-Q, reaching 44.04 F1 and outperforming \texttt{gate\_proj} (40.86) and \texttt{v\_proj} (38.91), while FFN modules are on average stronger than the query, key, and value projections. Overall, these results suggest that HAPS gains most from a sufficiently strong router backbone and a well-chosen injection site, with \texttt{o\_proj} working best among all final-layer modules we tested.

%% file: sections/exp/scaling.tex
\subsection{Scaling with Candidate Pool Size (RQ4)}
\label{sec:scaling}

We next study how performance and cost change as the candidate pool grows. For $K{=}5$, the pool contains Mistral-7B-Instruct-v0.3, Llama-3.1-8B-Instruct, Qwen2.5-7B-Instruct, Vicuna-7B-v1.5, and DeepSeek-LLM-7B-Chat; $K{=}6$ additionally includes Yi-1.5-6B-Chat. Since the $\sim$1B router backbone is shared, each extra candidate adds only a 17.4M-parameter low-level head (0.6M MLP + 16.8M projector), about 1.7\% of the backbone.
Table~\ref{tab:scaling_efficiency} shows that baselines without parameter adaptation degrade as $K$ increases, whereas HAPS improves from 37.14 to 41.41 F1 (+4.27 points). In particular, RouteLLM and GraphRouter lose 6.04 and 1.37 F1 points when the sixth candidate is added, while HAPS gains 4.27 points, suggesting that HAPS can exploit the added candidate through parameter adaptation. HAPS requires more training time than the baselines, mainly because its RL stage samples multi-agent interactions online; this stage uses only 300 training and 100 validation queries with early stopping (Appendix~\ref{app:data_collection}). On $K{=}5$, HAPS routes 1,000 queries in 25.60s, faster than RouteLLM (52.24s) though slower than GraphRouter (14.71s). Appendix~\ref{app:efficiency} details how LoRA generation and injection are vectorized and batched across requests at inference.

\begin{table*}[t]
\centering
\small
\renewcommand{\arraystretch}{1.1}
\setlength{\tabcolsep}{5pt}
\caption{
\textbf{Scalability and efficiency on larger candidate pools (HotpotQA).}
HAPS training time is reported as \emph{Warm-up + RL}; inference latency is measured on 1,000 queries with $K{=}5$.
Training time excludes the offline data sampling shared by all methods. Marking and units follow Table~\ref{tab:main_results}.
}
\label{tab:scaling_efficiency}
\begin{tabular*}{\textwidth}{@{\extracolsep{\fill}}lcccccc@{}}
\toprule
\multirow{2}{*}{\textbf{Method}} &
\multicolumn{3}{c}{\textbf{Scalability}} &
\multicolumn{2}{c}{\textbf{Training Time}} &
\multicolumn{1}{c}{\textbf{Inference}} \\
\cmidrule(lr){2-4}\cmidrule(lr){5-6}\cmidrule(lr){7-7}
& \textbf{F1@K=5} & \textbf{F1@K=6} & \textbf{$\Delta$} & \textbf{K=5} & \textbf{K=6} & \textbf{K=5 (1k)} \\
\midrule
Random & 23.28 & 19.43 & \textcolor{red!70!black}{-3.85} & -- & -- & -- \\
RouteLLM & 34.09 & 28.05 & \textcolor{red!70!black}{-6.04} & 6m 36s & 6m 32s & 52.24s \\
GraphRouter & \underline{35.35} & \underline{33.98} & \textcolor{red!70!black}{\underline{-1.37}} & 6m 04s & 7m 33s & 14.71s \\
\textbf{HAPS} & \textbf{37.14} & \textbf{41.41} & \textbf{\textcolor{green!50!black}{+4.27}} & 5m 19s + 32m 28s & 4m 21s + 21m 37s & 25.60s \\
\bottomrule
\end{tabular*}
\end{table*}

%% file: sections/exp/tradeoff.tex
\subsection{Performance \& Cost Trade-off (RQ5)}
\label{sec:tradeoff_exp}
\begin{table*}[t]
\centering
\small
\renewcommand{\arraystretch}{1.1}
\setlength{\aboverulesep}{0pt}
\setlength{\belowrulesep}{0pt}
\setlength{\tabcolsep}{2pt}

\caption{
\textbf{Performance--Cost Trade-off on HotpotQA.}
Reward is $\alpha \cdot \text{Perf} - \beta \cdot \text{Cost}_{\text{norm}}$, where Perf is the F1 score and Cost is the normalized cost.
All Weak and All Strong assign Llama-3.1-8B-Instruct and Qwen2.5-14B-Instruct, respectively, to both agents.
Marking follows Table~\ref{tab:main_results}.
}
\label{tab:hotpotqa_tradeoff}

\begin{tabularx}{\textwidth}{l YYY p{8pt} YYY p{8pt} YYY}
\toprule
\multirow{2}{*}{\textbf{Method}} &
\multicolumn{3}{c}{\textbf{$\alpha=0.8,\ \beta=0.2$}} & &
\multicolumn{3}{c}{\textbf{$\alpha=0.5,\ \beta=0.5$}} & &
\multicolumn{3}{c}{\textbf{$\alpha=0.2,\ \beta=0.8$}} \\
\cmidrule(lr){2-4}\cmidrule(lr){6-8}\cmidrule(lr){10-12}

& \multicolumn{1}{c}{\mbox{\textbf{Perf}~$\uparrow$}}
& \multicolumn{1}{c}{\mbox{\textbf{Cost}~$\downarrow$}}
& \multicolumn{1}{c}{\mbox{\textbf{Reward}~$\uparrow$}}
&
& \multicolumn{1}{c}{\mbox{\textbf{Perf}~$\uparrow$}}
& \multicolumn{1}{c}{\mbox{\textbf{Cost}~$\downarrow$}}
& \multicolumn{1}{c}{\mbox{\textbf{Reward}~$\uparrow$}}
&
& \multicolumn{1}{c}{\mbox{\textbf{Perf}~$\uparrow$}}
& \multicolumn{1}{c}{\mbox{\textbf{Cost}~$\downarrow$}}
& \multicolumn{1}{c}{\mbox{\textbf{Reward}~$\uparrow$}} \\
\midrule

\rowcolor{headerGray}
\multicolumn{12}{l}{\textit{\textbf{Static}}} \\
All Weak
& 0.3458 & \underline{0.0923} & 0.2582 & &
0.3458 & 0.0923 & 0.1267 & &
0.3458 & 0.0923 & -0.0047 \\
All Strong
& 0.3697 & 0.3362 & 0.2285 & &
\underline{0.3697} & 0.3362 & 0.0168 & &
\underline{0.3697} & 0.3362 & -0.1950 \\
\midrule

\rowcolor{headerGray}
\multicolumn{12}{l}{\textit{\textbf{Baselines}}} \\
Random
& 0.3433 & 0.2160 & 0.2314 & &
0.3433 & 0.2160 & 0.0636 & &
0.3433 & 0.2160 & -0.1042 \\
RouteLLM
& \textbf{0.4335} & 0.2111 & 0.3046 & &
0.3558 & 0.0939 & 0.1310 & &
0.3458 & 0.0923 & -0.0047 \\
GraphRouter
& 0.3458 & \underline{0.0923} & 0.2582 & &
0.3458 & 0.0923 & 0.1267 & &
0.3458 & 0.0923 & -0.0047 \\
IRT-Router
& \underline{0.4299} & 0.1377 & \underline{0.3164} & &
{0.3604} & \underline{0.0921} & \underline{0.1341} & &
{0.3552} & \underline{0.0881} & \underline{0.0005} \\
\midrule

{HAPS}
& {0.4194}
& \textbf{0.0886}
& \textbf{0.3178}
& &
\textbf{0.3709}
& \textbf{0.0857}
& \textbf{0.1426}
& &
\textbf{0.4181}
& \textbf{0.0861}
& \textbf{0.0148} \\
\bottomrule
\end{tabularx}
\end{table*}


In practical LLM routing, candidate models often differ in size, and larger models are typically more capable but also more expensive, so an explicit performance--cost trade-off is necessary.
We evaluate HAPS on HotpotQA with a smaller Llama-3.1-8B-Instruct and a larger Qwen2.5-14B-Instruct, pricing each Teacher--Student assignment by a simulated three-tier scheme that reflects relative cost levels rather than real API prices: \$0.2 per 1M input or output tokens when both agents use Llama, \$0.8 when both use Qwen, and \$0.5 when the two are mixed.
Following prior work~\citep{song2025irt, feng2024graphrouter}, we incorporate cost using a weighted reward
$R {=} \alpha {\cdot} \text{Perf} {-} \beta {\cdot} \text{Cost}_{\text{norm}}$,
where $\text{Perf}$ is task performance and $\text{Cost}_{\text{norm}}$ is the min-max normalized cost under this simulated scheme.
We report three regimes: \textbf{Performance-First} ($\alpha{=}0.8,\beta{=}0.2$), \textbf{Balanced} ($\alpha{=}0.5,\beta{=}0.5$), and \textbf{Cost-First} ($\alpha{=}0.2,\beta{=}0.8$). 
As shown in Table~\ref{tab:hotpotqa_tradeoff}, HAPS attains the highest reward and lowest cost in all three regimes. Even in the Performance-First setting, it remains competitive while reducing cost.
In the Cost-First setting, most baselines approach the All Weak strategy at around 35 F1, while HAPS still reaches 41.81 F1.
Notably, HAPS costs even less than All Weak, which routes both agents to the cheaper model, indicating that part of the saving comes from parameter adaptation shortening generations rather than from model selection alone.
Further details are provided in Appendix~\ref{app:reward_details}.

%% file: sections/exp/mixed.tex
\subsection{Routing Across Mixed Open- and Closed-Source LLMs (RQ6)}
\label{sec:mixed}

\begin{wrapfigure}{r}{0.50\columnwidth}
\vspace{-1.5\baselineskip}
  \centering
  \includegraphics[width=0.50\columnwidth]{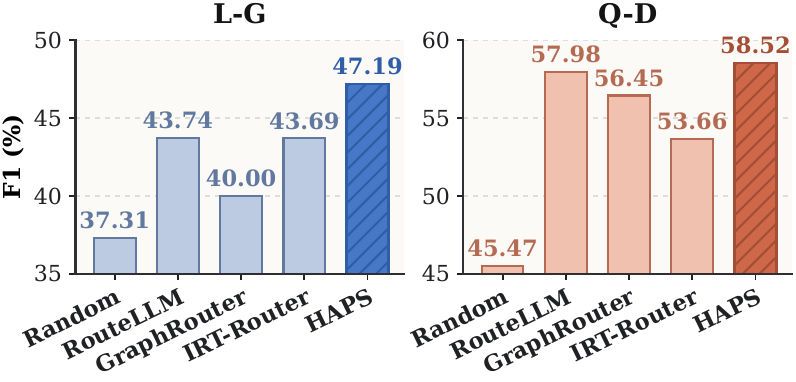}
\vspace{-1.3\baselineskip}
\caption{
\textbf{Mixed-source routing.}
HAPS on HotpotQA with open-source models and fixed proprietary APIs; generated parameters are applied only to the open-source model of each pair.
}
\label{fig:mixed_sources}
\vspace{-1.0\baselineskip}
\end{wrapfigure}


To validate the extensibility discussed in Remark~\ref{remark:mix} (Equation~(\ref{our-loss-imp})), we evaluate HAPS on two mixed-source pairs on HotpotQA: \textbf{L-G} (Llama-3.1-8B-Instruct \& GPT-4.1~Nano) and \textbf{Q-D} (Qwen2.5-7B-Instruct \& DeepSeek~V3).
As shown in Figure~\ref{fig:mixed_sources}, HAPS achieves the best F1 on both pairs. On L-G, it reaches 47.19 F1, outperforming the strongest baseline, RouteLLM, by 3.45 points. Since the API parameters are fixed, HAPS improves the system only through routing and by adapting the open-source model to complement the API.
On Q-D, HAPS also achieves the highest F1 of 58.52. Here the much stronger DeepSeek~V3 handles most queries, so the open-source Qwen is selected less often and leaves less room for parameter adaptation. HAPS still outperforms all baselines, showing that it remains effective even when the open-source model only supplements a powerful API. In both pairs, closed-source models are excluded from parameter generation through the indicator in Equation~(\ref{our-loss-imp}), so the same training procedure applies without modification.

%% file: sections/related.tex
\section{Related Work}
\textbf{LLM Routing.} LLM routing aims to optimize the performance--cost trade-off via dynamic model selection. Existing approaches are typically categorized into cascade systems, which invoke models sequentially based on estimated response quality~\citep{chen2023frugalgpt,dekoninck2024unified}, and predictive routers that directly map queries to suitable models~\citep{ding2024hybrid} using preference data~\citep{ong2024routellm}, embeddings~\citep{mohammadshahi2024routoo,chen2024routerdc,jitkrittum2025universal}, graphs~\citep{feng2024graphrouter}, or item response theory~\citep{song2025irt}, and recent work further assigns LLMs to agent roles in multi-agent systems~\citep{yue2025masrouter}. Together with supporting work on serving~\citep{stripelis2024tensoropera} and benchmarking~\citep{hu2024routerbench,huang2025routereval}, these methods treat candidate LLMs as \emph{fixed} and are limited to discrete architecture selection. HAPS bridges this gap by extending the decision space to jointly searchable parameter configurations.

\textbf{Dynamic Parameter Adaptation.} Parameter-Efficient Fine-Tuning (PEFT) is the de facto approach for adapting LLMs, spanning adapters, soft prompts, and LoRA~\citep{houlsby2019parameter,pfeiffer2020adapterhub,li2021prefix,lester2021power,hu2022lora,dettmers2023qlora}. However, conventional PEFT learns \emph{fixed} dataset-level parameters, limiting flexibility. To enable input-conditioned adaptation, prior work uses hypernetworks~\citep{ha2016hypernetworks} to generate parameters from inputs, improving transfer across tasks and domains~\citep{mahabadi2021parameter,ivison2023hint}. This extends naturally to LoRA, generating low-rank updates for cross-task generalization~\citep{lv2024hyperlora,charakorn2025text}. HAPS brings this paradigm to routing: rather than adapting a single fixed model, it generates LoRA parameters for whichever candidate is selected, so that model selection and parameter adaptation are optimized jointly under a single shared reward signal.

%% file: sections/limitations.tex
\section{Limitations}
\label{sec:limitations}
HAPS is primarily evaluated in two-agent settings with small candidate pools of up to six models; scaling to larger multi-agent workflows or more heterogeneous pools may require further routing and batching optimization.
Parameter generation also requires white-box access, so for closed-source APIs HAPS can only adjust routing, not behavior.
In addition, the RL stage samples multi-agent interactions online, which makes training slower than for routers trained purely offline; more sample-efficient optimization could reduce this cost.
Finally, we study only a subset of LoRA injection designs; future work could explore deeper or finer-grained parameter generation, such as injecting adapters into all layers or assigning layer-specific target modules to each candidate model.

%% file: sections/conclusion.tex
\section{Conclusion}
We presented HAPS, a hierarchical routing framework that jointly selects LLM architectures and generates input-conditioned LoRA parameters for the selected candidates.
Rather than treating each candidate as a fixed point in behavior space, HAPS lifts it to a region of reachable behaviors and routes over these regions.
Across HotpotQA and MMLU, multiple candidate pairs, larger candidate pools, cost-aware rewards, and mixed open- and closed-source deployments, HAPS consistently outperforms strong routing baselines.
We hope this perspective encourages a shift in LLM routing from routing among fixed models to routing over their reachable behaviors.

%% file: sections/acknowledgments.tex
\begin{ack}
The authors report no third-party funding or support for this work and no competing interests.
\end{ack}

%% file: sections/appendix/architecture.tex
\section{Architectural Details of HAPS}
\label{app:arch_details}
In this section, we provide a detailed specification of the high-level and low-level routing components within our hierarchical framework, complementing the overview in Section~\ref{sec:haps}.

\subsection{High-Level Router}
To minimize inference overhead, we employ a lightweight instruction-tuned LLM ({Llama-3.2-1B-Instruct}) as the high-level router $\mathbf{r}$, which reads the natural language routing prompt and expresses its decision as a JSON object assigning one candidate model to each agent in the system.

Since the set of valid assignments is finite, the router defines a distribution over them: during RL, assignments are sampled from this distribution, and at inference we select the most probable one. The chosen assignment is parsed into the vector $\mathbf{z}=\{z_i\}_{i=0}^{N-1}$, which dictates the model selection for each agent in the system and determines which low-level generators are invoked to produce the corresponding input-conditioned LoRA parameters for the selected candidates.

\subsection{Low-Level Router}
The role of the low-level router is to generate suitable LoRA parameters for the base models selected by the high-level router. We formulate this as a hypernetwork-based parameter generation task. Since candidate models in the pool possess distinct architectures (e.g., varying hidden state sizes), we instantiate a dedicated low-level routing head $\mathbf{s}_k$ for each candidate base model $m_k$.

Structurally, each low-level router $\mathbf{s}_k$ shares the backbone of the high-level router $\mathbf{r}$ (Section~\ref{sec:low_level}).
The routing prompt is encoded by this backbone, and its final-layer hidden states are mean-pooled over non-padding tokens.
On top of this pooled representation, $\mathbf{s}_k$ applies a {2-layer MLP} with 256 hidden units and ReLU activations.
This is subsequently followed by two parallel linear heads, which are responsible for independently projecting the latent vector into the LoRA rank-decomposition matrices $A$ and $B$, whose shapes depend on the hidden size of the target candidate.

Formally, for a selected base model $m_{z_i}$ with a hidden dimension of $d_{z_i}$, these heads generate $A \in \mathbb{R}^{d_{z_i} \times r}$ and $B \in \mathbb{R}^{r \times d_{z_i}}$, where the rank $r$ is a hyperparameter set to 8 based on empirical findings.
These generated parameters are injected into the output projection layer (\texttt{o\_proj}) of the final Transformer block in $m_{z_i}$, yielding the effective weight
\begin{equation}
    \tilde{W}^o_{z_i} = W^o_{z_i} + \alpha_{\mathrm{lora}} AB,
\end{equation}
where $W^o_{z_i}$ denotes the frozen weights of the final \texttt{o\_proj} layer, and $\alpha_{\mathrm{lora}}$ is the scaling factor.
This equation is only an equivalent form: in practice the update is never merged into $W^o_{z_i}$, and the adapter term is computed separately in batch to keep the per-query overhead low (Appendix~\ref{app:efficiency}).

%% file: sections/appendix/mas_setup.tex
\section{Detailed Multi-Agent System Configuration}
\label{app:mas_setup}
\begin{algorithm}[!tp]
    \SetAlgoLined
    \DontPrintSemicolon
    
    \KwIn{Question $q$, Budget $B$, Knowledge $\mathcal{I}$} 
    \KwOut{Final Answer $a$}
    
    $H_S, H_T \leftarrow \varnothing, \varnothing$\;
    
    \For{$t \gets 1$ \KwTo $B+1$}{
        \textcolor{gray}{\textbf{// Student Phase: Reasoning}}\;
        $s_{\mathrm{obs}} \leftarrow \mathrm{Observe}(t, H_S, H_T)$\;
        $s_{\mathrm{thought}} \leftarrow \mathrm{Think}(q, s_{\mathrm{obs}})$\;
        $s_{\mathrm{act}} \leftarrow \mathrm{Action}(q, s_{\mathrm{obs}}, s_{\mathrm{thought}})$\;
        
        \textcolor{gray}{\textbf{// Execution \& Termination}}\;
        \If{$s_{\mathrm{act}} \in \mathrm{Search}$}{
            $s_{\mathrm{res}} \leftarrow \mathrm{Retrieve}(s_{\mathrm{act}}, \mathcal{I})$\;
            $H_S \leftarrow H_S \oplus [(s_{\mathrm{act}}, s_{\mathrm{res}})]$\;
        }
        \ElseIf{$s_{\mathrm{act}} = \mathrm{Finish}[a]$}{
            \Return{$a$}\;
        }
        
        \textcolor{gray}{\textbf{// Teacher Phase: Feedback}}\;
        $t_{\mathrm{obs}} \leftarrow \mathrm{Observe}(t, H_S, H_T)$\;
        $t_{\mathrm{thought}} \leftarrow \mathrm{Think}(q, t_{\mathrm{obs}})$\;
        $t_{\mathrm{act}} \leftarrow \mathrm{Action}(q, t_{\mathrm{obs}}, t_{\mathrm{thought}})$\;
        $H_T \leftarrow H_T \oplus [(t_{\mathrm{act}}, t_{\mathrm{thought}})]$\;
    }
    \Return{$\mathrm{Null}$}
    
    \caption{HotpotQA Pipeline}
    \label{alg:hotpotqa}
\end{algorithm}
\begin{algorithm}[!tp]
    
    \SetAlgoLined
    \DontPrintSemicolon
    \KwIn{Problem $p$, Agents $S, T$, Limit $K_{\mathrm{round}}$}
    \KwOut{Final Answer $a$}
    
    $S_m, T_m \leftarrow [\;], [\;]$\;
    
    \For{$t \gets 1$ \KwTo $K_{\mathrm{round}}$}{
        \textcolor{gray}{\textbf{// Student Phase: Reasoning}}\;
        $s_{\mathrm{obs}} \leftarrow S.\mathrm{Observe}(t, S_m, T_m)$\;
        $s_{\mathrm{ans}} \leftarrow S.\mathrm{Answer}(p, s_{\mathrm{obs}})$\;
        $S_m.\mathrm{append}(s_{\mathrm{ans}})$\;
        
        \textcolor{gray}{\textbf{// Teacher Phase: Feedback}}\;
        $t_{\mathrm{obs}} \leftarrow T.\mathrm{Observe}(t, S_m)$\;
        
        $t_{\mathrm{feed}}, \epsilon \leftarrow T.\mathrm{Feedback}(p, t_{\mathrm{obs}})$\;
        
        \If{$\epsilon = \mathrm{True}$}{
            \Return{$s_{\mathrm{ans}}$}\;
        }
        $T_m.\mathrm{append}(t_{\mathrm{feed}})$\;
    }
    
    \Return{Last item of $S_m$}\;
    \caption{MMLU Pipeline}
    \label{alg:mmlu}
\end{algorithm}

Following the ReAct paradigm~\citep{yao2022react}, both benchmarks use a two-agent Teacher--Student system with minor task-specific adaptations in the action space and termination rule. All agent and router prompts used on both benchmarks are listed verbatim in Appendix~\ref{app:prompts}.
\paragraph{HotpotQA Protocol.}

For the HotpotQA dataset, which necessitates multi-hop reasoning, the system is equipped with an external retrieval interface. The interaction protocol is formalized as follows:

The Student agent operates with an action space consisting of two discrete primitives:
\begin{itemize}[leftmargin=*, topsep=0pt, parsep=0pt, itemsep=0pt]
    \item \texttt{Search[Entity]}: Retrieves the single paragraph most similar to \texttt{Entity} among those provided with the question, using an unsupervised SimCSE (RoBERTa-base) encoder.
    \item \texttt{Finish[Answer]}: Terminates the episode and submits the final \texttt{Answer} for evaluation.
\end{itemize}

The Teacher provides feedback by selecting \texttt{Continue} (validating the utility of the previous retrieval) or \texttt{Rethink} (prompting a revised search strategy with accompanying justification).

Both agents generate a reasoning trace before outputting their action.
Each agent also maintains a running summary of the interaction, which it rewrites every round to incorporate the newly acquired information.
Each episode allows a search budget of $B$ turns (set to 3 in our experiments), after which the Student must submit an answer in the final, $(B{+}1)$-th round.

To compute the reward, we normalize both the predicted answer $\hat{a}$ and the ground truth $g$ following standard protocols: (i) lower-casing; (ii) punctuation removal; (iii) article stripping (i.e., removing ``a'', ``an'', ``the''); and (iv) whitespace normalization.
We then compute the token-level F1 between the normalized strings as the task score $s \in [0, 1]$, which serves as the reward $R(\hat{y}, y)$ in Equation~(\ref{eq:rl-loss-agent}) after the shaping in Appendix~\ref{app:training}.
Algorithm~\ref{alg:hotpotqa} details the complete pipeline.

\paragraph{MMLU Protocol.}
Unlike HotpotQA, the MMLU {Student} uses \emph{no} external retrieval tools.  Each episode proceeds for at most
\(K_{\mathrm{round}}\) interaction rounds (default \(K_{\mathrm{round}} = 3\)).
At every round, the Student observes the Teacher's feedback from the previous
  round and generates a revised solution with answer. The {Teacher} reviews the solution, produces feedback and chooses between two discrete actions: \texttt{Continue} (requesting another revision) or \texttt{Submit} (terminating the episode and submitting the current solution as the final answer). If \texttt{Submit} is issued or $K_{\mathrm{round}}$ rounds are reached, the episode ends and the last solution and answer produced by the {Student} are evaluated.

The predicted answer is a single option in $\{\mathrm{A}, \mathrm{B}, \mathrm{C}, \mathrm{D}\}$ and is correct only if it exactly matches the ground-truth label, so the task score is binary, $s \in \{0, 1\}$, and is shaped into the reward as above. Algorithm~\ref{alg:mmlu} details the complete pipeline, including the Teacher's early-submission rule.

%% file: sections/appendix/data.tex
\section{Data Collection \& Implementation}\label{app:data_collection}

\paragraph{Benchmarks and Splits.}
For HotpotQA, we sample $3{,}000$ training ($\mathcal{D}_{\mathrm{train}}$), $1{,}000$ validation ($\mathcal{D}_{\mathrm{valid}}$), and $100$ test ($\mathcal{D}_{\mathrm{test}}$) questions from the \texttt{hotpot\_dev\_distractor\_v1.json} split.
For MMLU, we apply stratified sampling over all subjects to maintain broad coverage, obtaining $3{,}000$ training, $1{,}000$ validation, and $290$ test questions, each with four answer options.

\paragraph{Action-Level Dataset Construction.}
With two agent roles (Teacher/Student) and two candidate LLMs, a routing decision for each instance can be represented as an assignment vector
$\mathbf{z}=(z_T, z_S)$, where each entry specifies the selected model for one agent.
We define the routing action space as the set of all feasible assignment vectors, denoted by $\mathcal{A}_{\mathrm{act}}$;
in our setting, $|\mathcal{A}_{\mathrm{act}}| = 4$.

To obtain full-information supervision signals, we exhaustively execute every routing decision $\mathbf{z} \in \mathcal{A}_{\mathrm{act}}$
for each problem $p$ in the train/validation splits, and record the task score $s(p,\mathbf{z})$ defined in Appendix~\ref{app:mas_setup}.
This yields an expanded \textit{action-level} dataset:
\begin{equation}
    \widetilde{\mathcal{D}}_{\mathrm{split}}
    = \{(p, \mathbf{z}, s) \mid p \in \mathcal{D}_{\mathrm{split}}, \mathbf{z} \in \mathcal{A}_{\mathrm{act}}\},
\end{equation}
where $\mathrm{split} \in \{\mathrm{train}, \mathrm{valid}\}$.
Thus, each expanded dataset is $|\mathcal{A}_{\mathrm{act}}|$ times larger than its problem-level counterpart, covering all four routing decisions of every problem.

\paragraph{SFT Initialization Dataset.}
We first construct a warm-up dataset $\mathcal{D}_{\mathrm{warm}}$ to supervise the high-level router by selecting, for each training problem,
the highest-reward routing decision under exhaustive evaluation:
$\mathcal{D}_{\mathrm{warm}} {=} \{(p, \mathbf{z}^{\star}) \mid p \in \mathcal{D}_{\mathrm{train}},\;
\mathbf{z}^{\star} \in \arg\max_{\mathbf{z} \in \mathcal{A}_{\mathrm{act}}} s(p,\mathbf{z})\}$.
Our code supports several strategies for handling tied optimal decisions and label imbalance, such as keeping all tied decisions or over- and undersampling, which can be chosen according to the training budget.

\paragraph{Joint RL Dataset.}
To simulate a data-efficient online adaptation setting for Phase~2, we randomly select $300$ problems from $\mathcal{D}_{\mathrm{train}}$ and $100$ from $\mathcal{D}_{\mathrm{valid}}$ for each benchmark, forming $\mathcal{D}^{\mathrm{RL}}_{\mathrm{train}}$ and $\mathcal{D}^{\mathrm{RL}}_{\mathrm{valid}}$, respectively; the latter is used for checkpoint selection and early stopping.
During Phase~2, $\mathcal{D}^{\mathrm{RL}}_{\mathrm{train}}$ is used to jointly optimize the high-level router $\mathbf{r}_{\boldsymbol{\theta}}$ and low-level generators $\mathbf{s}_{\boldsymbol{\Gamma}}$.
The held-out $\mathcal{D}_{\mathrm{test}}$ is strictly reserved for final evaluation of the complete routing framework.

\paragraph{Baseline Adaptations.}
To align all the baselines with our main experimental protocol, which evaluates routing decisions solely by performance, we keep their original architectures and training procedures unchanged, and introduce only the minimal modifications required by our two-agent routing setup. For the performance--cost trade-off in Section~\ref{sec:tradeoff_exp}, we instead enable each baseline's native cost-aware routing.
\textbf{RouteLLM.}
Since RouteLLM makes binary decisions, we train an independent router for each agent role. Each router ignores cost and fixes the win-probability threshold at $\tau{=}0.5$ (other hyperparameters unchanged), selecting the strong model only when its predicted win probability is at least $0.5$.
\textbf{GraphRouter.}
GraphRouter builds a heterogeneous task--query--LLM graph with textual model descriptors; we instantiate descriptors for our two candidates and use the ``Performance First'' setting to obtain routing utilities.
\textbf{IRT-Router.}
We adopt the \emph{MIRT} variant and set $\beta{=}0$ (thus $\alpha{=}1$) to remove cost from its objective; we enable cold-start warm-up and use $\lambda{=}0.3$ following the sensitivity analysis in the original IRT-Router paper.

%% file: sections/appendix/training.tex
\section{Training \& Evaluation Details}

\label{app:training}
\begin{algorithm}[!tp]
    \SetAlgoLined
    \DontPrintSemicolon
    \SetNoFillComment
    \KwIn{Warm-Up Data $\mathcal{D}_{\mathrm{warm}}$; RL Data $\mathcal{D}^{\mathrm{RL}}_{\mathrm{train}}, \mathcal{D}^{\mathrm{RL}}_{\mathrm{valid}}$; Test Data $\mathcal{D}_{\mathrm{test}}$}
    \KwIn{Model Pool $\mathcal{M}$; Router $\mathbf{r}_{\boldsymbol{\theta}}$; Generators $\{\mathbf{s}_{\boldsymbol{\gamma}_m}\}_{m \in \mathcal{M}}$}
    \KwIn{LRs $\eta_{\mathrm{sft}}, \eta_{\theta}, \eta_{\gamma}$; Scale $\alpha_{\mathrm{lora}}$; Epochs $E_{\mathrm{sft}}, E_{\mathrm{rl}}$; Eval interval $K_{\mathrm{eval}}$; Patience $P$}
    \KwOut{Optimized Parameters $\boldsymbol{\theta}^*, \boldsymbol{\Gamma}^*$ and test score $\mathrm{score}_{\mathrm{test}}$}

    Initialize parameters $\boldsymbol{\theta}, \boldsymbol{\Gamma} = \{\boldsymbol{\gamma}_m\}_{m \in \mathcal{M}}$\;
    $\mathrm{score}_{\mathrm{best}} \leftarrow -\infty$;\quad $\mathrm{step} \leftarrow 0$;\quad $\mathrm{stall} \leftarrow 0$\;

    \textcolor{gray}{\textbf{// Phase 1: Supervised Warm-Up (SFT)}}\;
    \For{$\mathrm{epoch} \gets 1$ \KwTo $E_{\mathrm{sft}}$}{
        \ForEach{$(p, \mathbf{z}^{\star}) \in \mathrm{Shuffle}(\mathcal{D}_{\mathrm{warm}})$}{
            $\mathbf{z}_{\mathrm{pred}} \leftarrow \mathbf{r}_{\boldsymbol{\theta}}(p)$\;
            $\mathcal{L}_{\mathrm{sft}} \leftarrow \mathrm{CrossEntropy}(\mathbf{z}_{\mathrm{pred}}, \mathbf{z}^{\star})$\;
            $\boldsymbol{\theta} \leftarrow \boldsymbol{\theta} - \eta_{\mathrm{sft}} \nabla_{\boldsymbol{\theta}} \mathcal{L}_{\mathrm{sft}}$\;
        }
    }

    \textcolor{gray}{\textbf{// Phase 2: Joint RL Optimization}}\;
    \For{$\mathrm{epoch} \gets 1$ \KwTo $E_{\mathrm{rl}}$}{
        \ForEach{batch $\mathcal{B} \subset \mathrm{Shuffle}(\mathcal{D}^{\mathrm{RL}}_{\mathrm{train}})$}{
            \ForEach{$p \in \mathcal{B}$}{
                \tcp{1. Hierarchical Routing}
                $(m_T, m_S) \leftarrow \mathbf{r}_{\boldsymbol{\theta}}(p)$\;
                \tcp{2. Adaptive Parameter Generation}
                $\mathbf{w}_T \leftarrow \mathbf{s}_{\boldsymbol{\gamma}_{m_T}}(p), \quad \mathbf{w}_S \leftarrow \mathbf{s}_{\boldsymbol{\gamma}_{m_S}}(p)$\;
                \tcp{3. Vectorized Injection \& Interaction}
                $\hat{\mathbf{y}}_p, \log \boldsymbol{\pi}_{T,p}, \log \boldsymbol{\pi}_{S,p} \leftarrow \mathrm{RunMAS}(p \mid m_T^{\mathbf{w}_T}, m_S^{\mathbf{w}_S})$\;
                $s_p \leftarrow \mathrm{TaskMetric}(\hat{\mathbf{y}}_p, \mathbf{y}^*_p)$\;
                \lIf{$s_p = 0$}{$\hat R_p \leftarrow -1$} \lElse{$\hat R_p \leftarrow s_p$}
            }
            \tcp{4. Optimization Step}
            $\mathcal{L}_{\mathrm{rl}} \leftarrow -\frac{1}{|\mathcal{B}|}\sum_{p \in \mathcal{B}} \hat R_p \cdot (\sum \log \boldsymbol{\pi}_{T,p} + \sum \log \boldsymbol{\pi}_{S,p})$\;
            $\boldsymbol{\theta} \leftarrow \boldsymbol{\theta} - \eta_{\theta} \nabla_{\boldsymbol{\theta}} \mathcal{L}_{\mathrm{rl}}$\;
            \ForEach{model $m$ selected in $\mathcal{B}$}{
                $\boldsymbol{\gamma}_{m} \leftarrow \boldsymbol{\gamma}_{m} - \eta_{\gamma} \nabla_{\boldsymbol{\gamma}_m} \mathcal{L}_{\mathrm{rl}}$\;
            }
            \tcp{5. Interval Validation \& Early Stopping}
            $\mathrm{step} \leftarrow \mathrm{step}+1$\;
            \If{$\mathrm{step} \bmod K_{\mathrm{eval}} = 0$}{
                $\mathrm{score}_{\mathrm{val}} \leftarrow \mathrm{EvaluateBucketed}(\mathcal{D}^{\mathrm{RL}}_{\mathrm{valid}}, \boldsymbol{\theta}, \boldsymbol{\Gamma}, \alpha_{\mathrm{lora}})$\;
                \eIf{$\mathrm{score}_{\mathrm{val}} > \mathrm{score}_{\mathrm{best}}$}{
                    $\mathrm{score}_{\mathrm{best}} \leftarrow \mathrm{score}_{\mathrm{val}}$;\quad Save checkpoint $\boldsymbol{\theta}^*, \boldsymbol{\Gamma}^*$;\quad $\mathrm{stall} \leftarrow 0$\;
                }{
                    $\mathrm{stall} \leftarrow \mathrm{stall}+1$\;
                }
                \lIf{$\mathrm{stall} \geq P$}{stop training}
            }
        }
    }

    \textcolor{gray}{\textbf{// Final Evaluation}}\;
    $\mathrm{score}_{\mathrm{test}} \leftarrow \mathrm{EvaluateBucketed}(\mathcal{D}_{\mathrm{test}}, \boldsymbol{\theta}^*, \boldsymbol{\Gamma}^*, \alpha_{\mathrm{lora}})$\;
    \caption{HAPS Training: SFT Warm-Up \& Joint RL}
    \label{alg:haps_training}
\end{algorithm}

Algorithm~\ref{alg:haps_training} summarizes our two-phase protocol:
(i) supervised warm-up (SFT) of the high-level router on $\mathcal{D}_{\mathrm{warm}}$, and
(ii) joint on-policy optimization on $\mathcal{D}^{\mathrm{RL}}_{\mathrm{train}}$ with periodic validation on $\mathcal{D}^{\mathrm{RL}}_{\mathrm{valid}}$ for checkpoint selection and early stopping; the selected checkpoint is evaluated once on $\mathcal{D}_{\mathrm{test}}$ after training.
All main HAPS training runs are conducted on NVIDIA A800 80GB GPUs.
Table~\ref{tab:scaling_efficiency} reports representative training time and inference latency measurements.
In Phase~2, for each problem prompt $p$, the high-level router selects a joint assignment $\mathbf{z}=(z_T, z_S)$ for the Teacher and Student,
and the corresponding low-level generators produce input-conditioned LoRA parameters for the selected base models.
We then run the multi-agent interaction with the LoRA-injected models, sampling tokens during training and decoding greedily at evaluation, and record the log-probability of every token generated by \emph{both} agents.
In implementation, the likelihood $\log P(\mathbf{z}, \mathbf{w}_{\mathbf{z}} \mid p)$ in Equation~(\ref{eq:rl-loss-agent}) is computed from these token log-probabilities, which depend on the generated parameters and, through the shared backbone, on $\theta$. The per-episode loss is
\begin{equation}
\label{eq:rl-loss-agent-app}
\mathcal{L}_{\mathrm{rl}}
=
-\hat R \sum_{t \in \mathcal{T}}
\log \pi_{\theta,\mathbf{\Gamma}}\!\left(a_t \mid c_t\right),
\end{equation}
which is averaged over a batch of episodes before each update.
Here, $\mathcal{T}$ is the index set of all decoding steps produced by both agents in an episode; $a_t$ is the token emitted at step $t$ by the active agent, and $c_t$ is its conditioning context, consisting of the problem prompt, the interaction history, any retrieved evidence available up to that step, and role indicators.
$\pi_{\theta,\mathbf{\Gamma}}(a_t \mid c_t)$ is the probability assigned to $a_t$ by the active agent's LLM after injecting the generated LoRA parameters, which depends on the high-level router parameters $\theta$ and the low-level generator parameters $\mathbf{\Gamma}$.
$\hat R$ is the scalar episode-level reward derived from the task score $s$ (Appendix~\ref{app:mas_setup}); we apply a shaping rule $\hat R=-1$ if $s=0$, and otherwise $\hat R=s$.
The hyperparameters specific to HAPS are the LoRA rank ($r{=}8$), scaling factor ($\alpha_{\mathrm{lora}}{=}0.01$), injection site (final-layer \texttt{o\_proj}), generator width (256 hidden units), and interaction budgets ($B{=}3$ search turns for HotpotQA and $K_{\mathrm{round}}{=}3$ rounds for MMLU).

%% file: sections/appendix/ifeval.tex
\section{Single-Agent Routing on Instruction Following}
\label{app:ifeval}
\begin{table}[!tp]
\centering
\small
\renewcommand{\arraystretch}{1.1}
\caption{
\textbf{Single-agent instruction-following results on IFEval.}
L-Q and M-Q are two-candidate pools; L-Q-M adds Mistral-7B-Instruct-v0.3 as a third candidate.
Values are mean $\pm$ standard deviation over seeds $\{42,43,44\}$, in percentages with ``\%'' omitted. Marking follows Table~\ref{tab:main_results}.
}
\label{tab:ifeval_results}
\begin{tabularx}{\textwidth}{l YYY}
\toprule
\textbf{Method} & \textbf{L-Q} & \textbf{M-Q} & \textbf{L-Q-M} \\
\midrule
Random & 45.87\scriptsize{$\pm$0.52} & 39.53\scriptsize{$\pm$1.05} & 42.93\scriptsize{$\pm$0.09} \\
RouteLLM & 46.20\scriptsize{$\pm$0.75} & 44.93\scriptsize{$\pm$0.52} & 46.87\scriptsize{$\pm$0.96} \\
GraphRouter & \underline{47.00}\scriptsize{$\pm$0.71} & 45.33\scriptsize{$\pm$0.19} & \underline{47.47}\scriptsize{$\pm$0.09} \\
IRT-Router & 46.40\scriptsize{$\pm$0.28} & \underline{45.80}\scriptsize{$\pm$0.16} & 47.00\scriptsize{$\pm$0.00} \\
\midrule
\textbf{HAPS} & \textbf{47.40}\scriptsize{$\pm$0.90} & \textbf{46.73}\scriptsize{$\pm$0.75} & \textbf{49.33}\scriptsize{$\pm$0.74} \\
\bottomrule
\end{tabularx}
\end{table}

To test HAPS beyond multi-agent coordination and reasoning tasks, we evaluate it in a \emph{single-agent} setting ($N{=}1$) on IFEval-style instruction following, using verifiable instructions from the RLVR-IFeval pool~\citep{lambert2024tulu} built on the IFEval constraint set~\citep{zhou2023instruction}. For each instruction, the high-level router selects one model to answer it directly and the low-level router generates LoRA parameters for that model; no Teacher--Student interaction takes place. The reward is binary and given by programmatic constraint checkers rather than a judge model.

We use $2{,}000$ instances for training, $500$ for validation, and $500$ for testing, and report the mean reward on the test split over three seeds. Beyond two-candidate pools, we also route over a three-candidate pool (\textbf{L-Q-M}) to see whether the benefit survives a larger action space. All methods share the same data splits, candidate pools, and constraint checkers for training and evaluation.

As shown in Table~\ref{tab:ifeval_results}, HAPS attains the highest mean in all three pools, outperforming the strongest baseline by 0.40, 0.93, and 1.86 points on L-Q, M-Q, and L-Q-M. The margin is largest on the three-candidate pool, where the error bars of HAPS and the runner-up do not overlap. The benefit of joint architecture-and-parameter routing thus carries over to single-agent instruction following.

%% file: sections/appendix/efficiency.tex
\section{Efficient Inference Implementation}
\label{app:efficiency}

This appendix describes how HAPS supports \emph{per-input} LoRA parameters while preserving GPU parallelism.
The key challenge is that naively applying different adapters to different samples can introduce per-sample loops (thus $O(b)$ serial latency).
We avoid this by (i) separating the shared base computation from the adapter computation, and
(ii) vectorizing the adapter term with batched GEMM/BMM kernels.
This design follows multi-adapter serving systems that batch LoRA computations across heterogeneous requests~\citep{sheng2023s,chen2024punica}.

\subsection{Vectorized Dynamic LoRA Injection}
\label{app:vec_injection}

Consider a linear layer with frozen base weight $W_0 \in \mathbb{R}^{d_{\text{out}} \times d_{\text{in}}}$.
LoRA adds a low-rank update
$\Delta W = AB$,
where $A \in \mathbb{R}^{d_{\text{out}} \times r}$ and $B \in \mathbb{R}^{r \times d_{\text{in}}}$ with $r \ll \min(d_{\text{in}}, d_{\text{out}})$.
Let the batched hidden states be $H^{\mathrm{in}} \in \mathbb{R}^{b \times L \times d_{\text{in}}}$.
The layer output becomes
\begin{equation}
\begin{split}
H^{\mathrm{out}}
&=
H^{\mathrm{in}} W_0^{\top}
+
\alpha_{\mathrm{lora}}\, H^{\mathrm{in}} (AB)^{\top} \\
&=
H^{\mathrm{in}} W_0^{\top}
+
\alpha_{\mathrm{lora}}\, (H^{\mathrm{in}} B^{\top}) A^{\top},
\end{split}
\end{equation}
where $\alpha_{\mathrm{lora}}$ is the LoRA scaling factor; the second form never materializes the $d_{\text{out}} \times d_{\text{in}}$ update $AB$.

In HAPS, each sample $i$ has its own $(A_i, B_i)$ produced on the fly by the low-level generator.
Instead of looping over samples, we stack per-sample adapters into batch tensors
$\mathcal{A} \in \mathbb{R}^{b \times d_{\text{out}} \times r}$ and
$\mathcal{B} \in \mathbb{R}^{b \times r \times d_{\text{in}}}$, where $^{\top}$ transposes the last two dimensions.
We then compute the adapter term via two batched matrix multiplications:
\begin{equation}
H_{\mathrm{lora}}
=\alpha_{\mathrm{lora}}\cdot
\mathrm{BMM}\!\big(\mathrm{BMM}(H^{\mathrm{in}}, \mathcal{B}^{\top}), \mathcal{A}^{\top}\big),
\end{equation}
which yields $H_{\mathrm{lora}}\in\mathbb{R}^{b\times L\times d_{\text{out}}}$ in parallel across the batch.
Intuitively, the hidden states of each sample $i$ are first projected to a rank-$r$ subspace by $B_i$ and then mapped back by $A_i$,
while the base term $H^{\mathrm{in}} W_0^{\top}$ remains a standard highly optimized GEMM shared by all samples.
Because the frozen weights are never swapped in memory, this avoids per-request weight materialization and retains the throughput benefits of batched execution even under per-input adapters.

\subsection{Request Bucketing for Low Latency}
\label{app:bucketing}

To further improve runtime efficiency, we bucket incoming queries by the routed base model.
The high-level router first predicts the base model identity for each request; requests assigned to the same base model are then grouped and executed as a sub-batch.
Within each bucket, the low-level generator produces all per-request adapter parameters in one forward pass, and subsequent decoding runs on a single loaded base model with vectorized dynamic injection.
This reduces repeated model loading so that the base-model forward pass dominates end-to-end latency, while the per-request adapter computation remains fully batched within each bucket and adds little overhead.

%% file: sections/appendix/tradeoff.tex
\section{Trade-off Analysis}
\label{app:reward_details}

\subsection{Experimental Setup and Cost Simulation}
Following the protocols established in prior studies~\citep{song2025irt, feng2024graphrouter}, we quantify the trade-off between effectiveness and efficiency under explicit performance--cost reward configurations.
For each episode, we define the raw inference cost of a query, summed over both agents and all interaction rounds, as:
\begin{equation}
C_{\mathrm{raw}}
=
\frac{L_{\mathrm{in}}P_{\mathrm{in}}
+L_{\mathrm{out}}P_{\mathrm{out}}}{10^6}.
\end{equation}
where $L_{\text{in}}$ and $L_{\text{out}}$ are the input and output token counts, and
$P_{\text{in}}$ and $P_{\text{out}}$ are the corresponding unit prices (per 1M tokens) under the three-tier pricing scheme used in Section~\ref{sec:tradeoff_exp}: \$0.2/\$0.2 for Llama+Llama, \$0.5/\$0.5 for mixed Llama+Qwen routing, and \$0.8/\$0.8 for Qwen+Qwen.

To make cost commensurate with the performance metric (e.g., $\text{Perf}\in[0,1]$ for F1), we apply min-max normalization using training-set statistics:
\begin{equation}
\text{Cost}_{\text{norm}}=
\frac{C_{\text{raw}}-C_{\min}}{C_{\max}-C_{\min}},
\end{equation}
where $C_{\min}$ and $C_{\max}$ are computed on the training split and then fixed for validation/test evaluation.

\paragraph{Reward Definition.}
We evaluate performance--cost trade-offs via a scalar reward:
\begin{equation}
\text{Reward} = \alpha \cdot \text{Perf} - \beta \cdot \text{Cost}_{\text{norm}},
\label{eq:tradeoff_reward}
\end{equation}
where $(\alpha,\beta)$ specifies the deployment preference and $\text{Perf}$ is the F1 score on HotpotQA. For each $(\alpha,\beta)$, HAPS is trained with Equation~(\ref{eq:tradeoff_reward}) as its RL reward, while the baselines use their native cost-aware routing under the same $(\alpha,\beta)$ weights, as described in Appendix~\ref{app:data_collection}.

The cost excludes LoRA generation and injection, whose overhead is small relative to the base-model forward pass under batched execution (Appendix~\ref{app:efficiency}), so it reflects token usage and model selection.

\subsection{Performance Analysis across Scenarios}
We evaluate HAPS under three configurations $(\alpha,\beta)$ that reflect different deployment priorities.

\paragraph{Performance-First ($\alpha{=}0.8,\ \beta{=}0.2$).}
In this regime, the system prioritizes accuracy.
As shown in Table~\ref{tab:hotpotqa_tradeoff}, RouteLLM attains 43.35 F1 but incurs substantial cost (0.2111) due to frequent selection of the expensive model.
In contrast, HAPS achieves a comparable 41.94 F1 while reducing the cost to 0.0886, a 58\% reduction relative to RouteLLM.
This suggests HAPS selectively leverages the cheaper model when it suffices, aided by input-conditioned parameter generation.

\paragraph{Balanced ($\alpha{=}0.5,\ \beta{=}0.5$).}
In the Balanced regime, HAPS achieves the highest aggregate reward (0.1426), outperforming the runner-up IRT-Router (0.1341).
These results suggest that coupling discrete model selection with continuous parameter adaptation achieves a better balance between performance and cost than routing among fixed candidate models alone.

\paragraph{Cost-First ($\alpha{=}0.2,\ \beta{=}0.8$).}
As discussed in Section~\ref{sec:tradeoff_exp}, HAPS costs less than All Weak under the same decoding setting (0.0861 vs.\ 0.0923) while maintaining 41.81 F1 (vs.\ 34.58).
Since no routing decision can be cheaper than assigning the cheaper model to both agents, the remaining saving must come from fewer generated tokens, e.g., through more concise trajectories.

%% file: sections/appendix/case_study.tex
\section{What the Generated Parameters Change}
\label{app:case_study}
\begin{figure}[!tp]
\centering
\small
\renewcommand{\arraystretch}{1.25}
\setlength{\tabcolsep}{5pt}
\newcommand{\qry}[1]{{\footnotesize\texttt{#1}}}
\newcommand{\tch}[1]{\par\textcolor{black!55}{\textit{Teacher:}}~{\footnotesize\texttt{#1}}}
\setlength{\aboverulesep}{0pt}
\setlength{\belowrulesep}{0pt}
\begin{tabularx}{\textwidth}{@{}c >{\raggedright\arraybackslash}X !{\color{black!30}\vrule} >{\raggedright\arraybackslash}X@{}}
\toprule
\multicolumn{3}{@{}p{\textwidth}@{}}{\textbf{Question.} \textit{What was the name of the lawyer who took the main role in the report that helped land Pete Rose a lifetime suspension from baseball?} \quad \textbf{Answer:} \texttt{John M.~Dowd}} \\
\midrule
\textbf{Round} &
\cellcolor{red!7}\textbf{Frozen base models} &
\cellcolor{green!8}\textbf{With generated parameters} \\
\midrule
1 &
\qry{Search[Pete Rose lifetime suspension lawyer]} $\rightarrow$ a television film about Pete Rose; no lawyer named
\tch{[Rethink]} &
\qry{Search[Pete Rose lifetime suspension lawyer]} $\rightarrow$ same paragraph
\tch{[Rethink]} \\
2 &
\qry{Search[Pete Rose lifetime suspension lawyer]} \newline \textcolor{red!70!black}{\textit{identical query repeated}} $\rightarrow$ same paragraph
\tch{[Rethink]} &
\qry{Search[lawyer in Pete Rose report]} \newline \textcolor{green!45!black}{\textit{query rephrased}} $\rightarrow$ \textit{``The Dowd Report \ldots\ was prepared by Special Counsel to the Commissioner, John M.~Dowd \ldots''}
\tch{[Continue]} \\
3 &
\qry{Search[Pete Rose lifetime suspension lawyer]} \newline \textcolor{red!70!black}{\textit{identical query repeated}} $\rightarrow$ same paragraph
\tch{[Rethink]} &
\qry{Finish[John M.~Dowd]} \\
4 &
\qry{Finish[Judge Kenesaw Mountain Landis]} \newline \textcolor{red!70!black}{\textit{name absent from all retrieved text}} &
-- \\
\midrule
\textbf{Final} &
\textcolor{red!70!black}{\textit{wrong answer}} \hfill F1 $=0.0$ &
\textcolor{green!45!black}{\texttt{John M.~Dowd}} (correct) \hfill F1 $=1.0$ \\
\bottomrule
\end{tabularx}
\caption{Case study on HotpotQA (M-Q pair) with the high-level routing decision held fixed, comparing the frozen base models with the same models after injecting the generated LoRA. Each cell shows the Student's action in that round, followed by the Teacher's response to it.}
  \label{fig:case_study}
\end{figure}

We examine how the generated parameters change agent behavior by holding the high-level decision fixed and toggling only the generated parameters: for each test question we take the architecture that HAPS itself selected, then run the same multi-step agent twice with greedy decoding, once on the frozen base models and once with the generated LoRA injected. Any difference is therefore attributable to the parameters alone. The example below is from the M-Q pair, where HAPS routes the Teacher to Mistral-7B-Instruct-v0.3 and the Student to Qwen2.5-7B-Instruct.

In the example of Figure~\ref{fig:case_study}, the first retrieval is uninformative: the query returns a paragraph about a television film on Pete Rose that does not name the lawyer. Both runs receive this paragraph in Round~1 and are told by the Teacher to rethink. Without the generated parameters, the Student repeats the identical query in every remaining round, receives the same paragraph each time, and finally answers with a name that appears in none of the retrieved text. With them, the Student rephrases the query, which retrieves the paragraph on the Dowd Report stating that it was prepared by John M.~Dowd; the Teacher confirms, and the Student answers correctly in the next round. The generated parameters thus change how the agents respond when a retrieval returns irrelevant evidence.

%% file: sections/appendix/prompts.tex
\section{Prompt Templates}
\label{app:prompts}
Figures~\ref{fig:prompt_first}--\ref{fig:prompt_last} list the prompts used in our experiments: the Thought \& Action and Context prompts of the HotpotQA agents, the prompts of the MMLU agents, and the system and user prompts that the high-level router receives on both benchmarks. Placeholders in braces are filled at run time.

\begin{figure}[!tp]
  \centering
  \begin{tcolorbox}[
      width=\textwidth,
      enhanced,
  colframe=black!55,
  colback=black!6,
  colbacktitle=black!65,
      coltitle=white,
      title=\Large \textbf{HotpotQA Teacher Thought \& Action Prompt},
      fonttitle=\bfseries,
      arc=4mm,
      boxsep=5pt,
      top=5pt, bottom=5pt,
  ]

You are the Teacher agent. Your task is to guide the Student agent to answer the question. You should analyze whether the student's step is logically helpful, provide an analysis, and give your final action. You can also give advice on the student's future step.\\
Context refers to the summary of previous interactions.\\
Question is the question you need to guide the Student to answer.\\
Observation is the last round of what you guided and what the Student did after you guided.\\
Thought should reason about the current situation and determine what action to take.\\
Action can be of two types:\\
(1) [Continue], which means the student made a good decision and should continue the process.\\
(2) [Rethink], which means the student's previous step is wrong and should consider another step.\\

Please note: \\
You can analyze the correctness of the Student agent's action, summarize the useful information the Student found, or provide suggestions for subsequent steps.\\
Your action step has only two types: [Continue] or [Rethink]. You can only take one of the above actions. Please be an encouraging teacher most of the time; that is, unless the student is completely wrong, use the [Continue] action more often.\\

[INPUT]\\
Context: \{context\}\\
Question: \{question\}\\
Observation: \{observation\}\\

[OUTPUT REQUIREMENTS]\\
Your output must strictly follow the format:\\

Thought: <Your thinking process on your action.>\\
Action: <Your action on the current situation. [Continue] or [Rethink]>\\

You must output exactly in the format, beginning with "Thought:" when you output your thought, and beginning with "Action:" when you output your action.\\
Notice that the action **MUST** be wrapped by brackets "[]".\\
Keep your entire output within 256 tokens.
\end{tcolorbox}
\caption{The Thought \& Action prompt of the Teacher agent on HotpotQA.}
\label{fig:prompt_first}
\end{figure}

\begin{figure}[!tp]
  \centering
  \begin{tcolorbox}[
      width=\textwidth,
      enhanced,
  colframe=black!55,
  colback=black!6,
  colbacktitle=black!65,
      coltitle=white,
      title=\Large \textbf{HotpotQA Teacher Context Prompt},
      fonttitle=\bfseries,
      arc=4mm,
      boxsep=5pt,
      top=5pt, bottom=5pt,
  ]

You are the Teacher agent, and your task is to guide the Student agent to answer the question.\\
Now you are provided with a previous summary, as well as new messages that were not included in the original summary. You need to make a new summary.\\
Your summary should encapsulate the main points of the new messages and integrate them into the existing summary to create a comprehensive recap. Highlight the key issues discussed, decisions made, and any actions assigned. Record the helpful factual information given by the search engine.\\

[INPUT]\\
Question: \{question\}\\
Previous Summary: \{context\}\\
New Observation: \{observation\}\\
New Thought: \{thought\}\\
New Action: \{action\}\\

[OUTPUT REQUIREMENTS]\\
Your output must strictly follow the format:\\

Summary: <Your new summary based on the information.>\\

You must output exactly in the format, beginning with "Summary:" when you output your new summary.\\
Keep your entire summary within 256 tokens.
\end{tcolorbox}
\caption{The Context prompt of the Teacher agent on HotpotQA.}
\end{figure}

\begin{figure}[!tp]
  \centering
  \begin{tcolorbox}[
      width=\textwidth,
      enhanced,
  colframe=black!55,
  colback=black!6,
  colbacktitle=black!65,
      coltitle=white,
      title=\Large \textbf{HotpotQA Student Thought \& Action Prompt},
      fonttitle=\bfseries,
      arc=4mm,
      boxsep=5pt,
      top=5pt, bottom=5pt,
  ]
You are the Student agent. Your task is to answer the question under the guidance of the Teacher agent and the given information. First, you should make a reasonable plan and choose the appropriate action.\\
Context refers to the summary of previous interactions.\\
Question is the question you need to find the answer to.\\
Observation is the last action you took and the guidance of the Teacher agent.\\
Budget refers whether you can search or must submit the answer.\\
Thought should reason about the current situation and determine what action to take.\\
Action can be of two types:\\
(1) Search[entity], which invokes a local searcher to provide relevant information about the entity.\\
(2) Finish[answer], which returns the final answer and completes the task.\\

Please note: \\
If you need more information to answer the question, output "Search[entity]". \\
If you find enough information to answer the question, output "Finish[answer]".\\
When submitting your answer, please try to submit the full answer if you think it is ambiguous, e.g. "movie director" is better than "director"! \\
The answer to the question should be as accurate and concise as possible, i.e. try not to answer the question with long sentences. Please answer the yes-no question with either "yes" or "no".\\
Be aware of your search time limit. If you are told in the "Budget" that you must answer the question, return the action "Finish[answer]".\\

[INPUT]\\
Context: \{context\}\\
Question: \{question\}\\
Observation: \{observation\}\\
Budget: \{budget\}\\

[OUTPUT REQUIREMENTS]\\
Your output must strictly follow the format:\\

Thought: <Your thinking process on your action.>\\
Action: <Your action on the current situation.>\\

You must output exactly in the format, beginning with "Thought:" when you output your thought, and beginning with "Action:" when you output your action.\\
Keep your entire output within 256 tokens.
\end{tcolorbox}
\caption{The Thought \& Action prompt of the Student agent on HotpotQA.}
\end{figure}

\begin{figure}[!tp]
  \centering
  \begin{tcolorbox}[
      width=\textwidth,
      enhanced,
  colframe=black!55,
  colback=black!6,
  colbacktitle=black!65,
      coltitle=white,
      title=\Large \textbf{HotpotQA Student Context Prompt},
      fonttitle=\bfseries,
      arc=4mm,
      boxsep=5pt,
      top=5pt, bottom=5pt,
  ]
You are the Student agent. Your task is to answer the question under the guidance of the Teacher agent.\\
Now you are provided with a previous summary, as well as new messages that were not included in the original summary. You need to make a new summary.\\
Your summary should encapsulate the main points of the new messages and integrate them into the existing summary to create a comprehensive recap. Highlight the key issues discussed, decisions made, and any actions assigned. Record the helpful factual information given by the search engine.\\

[INPUT]\\
Question: \{question\}\\
Previous Summary: \{context\}\\
New Observation: \{observation\}\\
New Thought: \{thought\}\\
New Action: \{action\}\\

[OUTPUT REQUIREMENTS]\\
Your output must strictly follow the format:\\

Summary: <Your new summary based on the information.>\\

You must output exactly in the format, beginning with "Summary:" when you output your new summary.\\
Keep your entire summary within 256 tokens.
\end{tcolorbox}
\caption{The Context prompt of the Student agent on HotpotQA.}
\end{figure}

\begin{figure}[!tp]
  \centering
  \begin{tcolorbox}[
      width=\textwidth,
      enhanced,
  colframe=black!55,
  colback=black!6,
  colbacktitle=black!65,
      coltitle=white,
      title=\Large \textbf{MMLU Teacher Prompt},
      fonttitle=\bfseries,
      arc=4mm,
      boxsep=5pt,
      top=5pt, bottom=5pt,
  ]
You are the Teacher agent. Your goal is to review the Student agent's current solution to a multiple-choice question and provide targeted feedback, and decide whether the solution is correct and ready for submission.\\

[INPUT]\\
Here is the multiple-choice question and the options: \\
\{question\}\\
Here is the Student agent's current choice and rationale: \\
\{current\_solution\}\\

[OUTPUT REQUIREMENTS]\\
Your output must strictly follow this format:\\

Feedback: <Your feedback on the Student's last round choice and rationale.>\\
Submit: <Decide whether the choice is correct to be submitted, 0 (not enough to submit) or 1 (enough to submit).>\\

You must output exactly in the format, beginning with "Feedback:" and "Submit:" when you output feedback and submit.\\
Keep your entire output within 512 tokens.
\end{tcolorbox}
\caption{The prompt of the Teacher agent on MMLU.}
\end{figure}

\begin{figure}[!tp]
  \centering
  \begin{tcolorbox}[
      width=\textwidth,
      enhanced,
  colframe=black!55,
  colback=black!6,
  colbacktitle=black!65,
      coltitle=white,
      title=\Large \textbf{MMLU Student Prompt},
      fonttitle=\bfseries,
      arc=4mm,
      boxsep=5pt,
      top=5pt, bottom=5pt,
  ]
You are the Student agent. Your task is to read a multiple-choice question and provide the most likely correct answer based on the options given, revising your previous answer according to the Teacher's feedback.\\

[INPUT]\\
Here is the multiple-choice question and options:\\
\{question\}\\
Here is your last round choice and rationale:\\
\{last\_solution\}\\
Here is the Teacher agent's feedback of your last round choice:\\
\{last\_feedback\}\\

[OUTPUT REQUIREMENTS]\\
Your output must strictly follow the format:\\

Rationale: <Your analysis process on the question, incorporating previous rationale and feedback.>\\
Choice: <Your final choice, A or B or C or D.>\\

You must output exactly in the format, beginning with "Rationale:" and "Choice:" when you output rationale and choice.\\
Keep your entire output within 512 tokens.
\end{tcolorbox}
\caption{The prompt of the Student agent on MMLU.}
\end{figure}

\begin{figure}[!tp]
  \centering
  \begin{tcolorbox}[
      width=\textwidth,
      enhanced,
  colframe=black!55,
  colback=black!6,
  colbacktitle=black!65,
      coltitle=white,
      title=\Large \textbf{HotpotQA \& MMLU System Prompt},
      fonttitle=\bfseries,
      arc=4mm,
      boxsep=5pt,
      top=5pt, bottom=5pt,
  ]
You are a router that must output only a single JSON object that matches the schema\\
\{"teacher": <id>, "student": <id>\}.\\
The integers must be valid model IDs listed in the prompt.\\
No explanation, no extra text. If unsure, still output a valid JSON object.
\end{tcolorbox}
\caption{The system prompt for routing on both HotpotQA and MMLU.}
\end{figure}

\begin{figure}[!tp]
  \centering
  \begin{tcolorbox}[
      width=\textwidth,
      enhanced,
  colframe=black!55,
  colback=black!6,
  colbacktitle=black!65,
      coltitle=white,
      title=\Large \textbf{HotpotQA User Prompt},
      fonttitle=\bfseries,
      arc=4mm,
      boxsep=5pt,
      top=5pt, bottom=5pt,
  ]
\textbf{\#\#\# Dataset Background\\}
The dataset we use is HotPotQA, where each data point includes a question and relevant information. We employ a multi-agent system comprising a Teacher and a Student agent to answer these questions. The agents alternate actions until either the search budget is exhausted or the Student provides a final answer.\\

\textbf{\#\#\# Student Agent Introduction\\}
The Student agent is responsible for searching for information and answering the question, guided by the Teacher agent. The Student's action space includes:\\
- Search[entity]: This action uses a retrieval tool to search for relevant information from the dataset, constructed using the SimCSE model (unsup-simcse-roberta-base).\\
- Finish[answer]: This action returns the final answer and concludes the task.\\

\textbf{\#\#\# Teacher Agent Introduction\\}
The Teacher agent evaluates and guides the Student agent by analyzing its steps and providing feedback. The Teacher can:\\
- [Continue]: Indicate that the Student's current step is correct and they should proceed.\\
- [Rethink]: Indicate that the Student's action is incorrect and they should reconsider their approach.\\

\textbf{\#\#\# Specific Question\\}
\{question\}\\

\textbf{\#\#\# Candidates\\}
0: \{model\_0\}\\
1: \{model\_1\}\\

\textbf{\#\#\# Task\\}
Assign a model to each agent to maximize performance.\\
Return exactly one JSON object on a single line:\\
\{\{"teacher": <id>, "student": <id>\}\}\\
where <id> is an integer, 0 or 1. Do not include any extra text.
\end{tcolorbox}
\caption{The user prompt for routing on HotpotQA.}
\end{figure}

\begin{figure}[!tp]
  \centering
  \begin{tcolorbox}[
      width=\textwidth,
      enhanced,
  colframe=black!55,
  colback=black!6,
  colbacktitle=black!65,
      coltitle=white,
      title=\Large \textbf{MMLU User Prompt},
      fonttitle=\bfseries,
      arc=4mm,
      boxsep=5pt,
      top=5pt, bottom=5pt,
  ]
\textbf{\#\#\# Dataset Background}

The dataset we use is MMLU, where each data point includes a multiple-choice question with the options.
We employ a multi-agent system comprising a Teacher and a Student agent to solve these questions.
The agents alternate actions until either round limit 3 is exhausted or the Teacher thinks the current choice of the Student is correct enough to be submitted.
\\

\textbf{\#\#\# Student Agent Introduction}

The Student agent is responsible for solving the question, guided by the Teacher agent.
It will be given the question with the options, its last round solution and Teacher's feedback on its last round solution. The Student needs to provide a new better solution.
\\

\textbf{\#\#\# Teacher Agent Introduction}

The Teacher agent evaluates and guides the Student agent by analyzing its current solution and providing feedback.
It will be given the question with the options and Student's current solution.
The Teacher needs to provide feedback on the solution.
\\

\textbf{\#\#\# Specific Question With the Options}

\{question\}
\\

\textbf{\#\#\# Candidates}

0: \{model\_0\}

1: \{model\_1\}
\\

\textbf{\#\#\# Task}

Assign a model to each agent to maximize performance.

Return exactly one JSON object on a single line:

\{\{"teacher": <id>, "student": <id>\}\}

where <id> is an integer, 0 or 1. Do not include any extra text.
\end{tcolorbox}
\caption{The user prompt for routing on MMLU.}
\label{fig:prompt_last}
\end{figure}

%% file: sections/appendix/broader_impacts.tex
\section{Broader Impacts}
\label{app:broader_impacts}
HAPS is intended to improve the effectiveness and efficiency of multi-model LLM deployment by routing queries to suitable models and adapting selected open-source models with lightweight generated parameters.
This may reduce unnecessary use of stronger or more expensive models and make adaptive routing systems more accessible.
At the same time, stronger automated LLM routing systems inherit the risks of their underlying models, including incorrect, biased, or harmful outputs when deployed without appropriate monitoring.
Practical deployments should therefore combine HAPS with task-specific safety checks, model-use policies, and human oversight, especially when used in sensitive domains. Since HAPS modifies model behavior through generated parameters, such checks should be applied to the adapted models rather than only to their frozen base versions.

%% file: checklist.tex
\section*{NeurIPS Paper Checklist}

\begin{enumerate}

\item {\bf Claims}
    \item[] Question: Do the main claims made in the abstract and introduction accurately reflect the paper's contributions and scope?
    \item[] Answer: \answerYes{}.
    \item[] Justification: The abstract and introduction clearly state the main contribution of HAPS: jointly routing over LLM architectures and input-conditioned parameter adaptations. They also summarize the hierarchical design, training objective, and experimental scope, which are supported by the main results and ablation studies in Section~4.

\item {\bf Limitations}
    \item[] Question: Does the paper discuss the limitations of the work performed by the authors?
    \item[] Answer: \answerYes{}.
    \item[] Justification: We include a dedicated Limitations section in Section~\ref{sec:limitations}.

\item {\bf Theory assumptions and proofs}
    \item[] Question: For each theoretical result, does the paper provide the full set of assumptions and a complete (and correct) proof?
    \item[] Answer: \answerNA{}.
    \item[] Justification: This paper proposes an empirical LLM routing framework and does not present any formal theoretical results, theorems, assumptions, or proofs.

    \item {\bf Experimental result reproducibility}
    \item[] Question: Does the paper fully disclose all the information needed to reproduce the main experimental results of the paper to the extent that it affects the main claims and/or conclusions of the paper (regardless of whether the code and data are provided or not)?
    \item[] Answer: \answerYes{}.
    \item[] Justification: We provide the information needed to reproduce the main results in Sections~\ref{pre}--\ref{MO} and Appendices~\ref{app:arch_details}--\ref{app:training} and~\ref{app:prompts}, including the model architecture, data construction, prompts, baseline adaptations, and the training and evaluation procedures used in all experiments.

\item {\bf Open access to data and code}
    \item[] Question: Does the paper provide open access to the data and code, with sufficient instructions to faithfully reproduce the main experimental results, as described in supplemental material?
    \item[] Answer: \answerYes{}.
    \item[] Justification: We release our code at the link given in the abstract, and describe the data, implementation details, prompts, and evaluation protocols in the appendix.

\item {\bf Experimental setting/details}
    \item[] Question: Does the paper specify all the training and test details (e.g., data splits, hyperparameters, how they were chosen, type of optimizer) necessary to understand the results?
    \item[] Answer: \answerYes{}.
    \item[] Justification: We specify the experimental setup in Section~4, provide additional details on datasets, model pairs, hyperparameters, training, and evaluation in Appendices~\ref{app:mas_setup}--\ref{app:training}, and list all agent and router prompts used on both benchmarks in Appendix~\ref{app:prompts}.

\item {\bf Experiment statistical significance}
    \item[] Question: Does the paper report error bars suitably and correctly defined or other appropriate information about the statistical significance of the experiments?
    \item[] Answer: \answerNo{}.
    \item[] Justification: We do not report error bars or formal significance tests because the main experiments require expensive multi-agent LLM inference. Following common practice in comparable LLM routing studies, we instead report results on a separate test set together with extensive ablations and diagnostic analyses; Appendix~\ref{app:ifeval} additionally reports the mean and standard deviation over three random seeds for the single-agent experiments.

\item {\bf Experiments compute resources}
    \item[] Question: For each experiment, does the paper provide sufficient information on the computer resources (type of compute workers, memory, time of execution) needed to reproduce the experiments?
    \item[] Answer: \answerYes{}.
    \item[] Justification: We report compute-related information in Appendix~\ref{app:training} and Table~\ref{tab:scaling_efficiency}, including the GPU type used for main training runs, representative training time, and inference latency.
    
\item {\bf Code of ethics}
    \item[] Question: Does the research conducted in the paper conform, in every respect, with the NeurIPS Code of Ethics \url{https://neurips.cc/public/EthicsGuidelines}?
    \item[] Answer: \answerYes{}.
    \item[] Justification: The research uses public benchmarks and existing LLMs, does not involve human subjects, and fully conforms to the NeurIPS Code of Ethics.

\item {\bf Broader impacts}
    \item[] Question: Does the paper discuss both potential positive societal impacts and negative societal impacts of the work performed?
    \item[] Answer: \answerYes{}.
    \item[] Justification: We discuss potential positive and negative societal impacts in Appendix~\ref{app:broader_impacts}.
    
\item {\bf Safeguards}
    \item[] Question: Does the paper describe safeguards that have been put in place for responsible release of data or models that have a high risk for misuse (e.g., pre-trained language models, image generators, or scraped datasets)?
    \item[] Answer: \answerNA{}.
    \item[] Justification: We do not release pretrained models or datasets that pose a high risk of misuse; the released code only builds on existing publicly available models and benchmarks.

\item {\bf Licenses for existing assets}
    \item[] Question: Are the creators or original owners of assets (e.g., code, data, models), used in the paper, properly credited and are the license and terms of use explicitly mentioned and properly respected?
    \item[] Answer: \answerYes{}.
    \item[] Justification: We cite them in the paper properly.

\item {\bf New assets}
    \item[] Question: Are new assets introduced in the paper well documented and is the documentation provided alongside the assets?
    \item[] Answer: \answerYes{}.
    \item[] Justification: The paper introduces code and experimental artifacts for HAPS; the method, prompts, training procedure, and evaluation protocol are documented in the paper and appendix, and the code is released in our public GitHub repository linked in the abstract.

\item {\bf Crowdsourcing and research with human subjects}
    \item[] Question: For crowdsourcing experiments and research with human subjects, does the paper include the full text of instructions given to participants and screenshots, if applicable, as well as details about compensation (if any)? 
    \item[] Answer: \answerNA{}.
    \item[] Justification: This work does not involve crowdsourcing or research with human subjects.

\item {\bf Institutional review board (IRB) approvals or equivalent for research with human subjects}
    \item[] Question: Does the paper describe potential risks incurred by study participants, whether such risks were disclosed to the subjects, and whether Institutional Review Board (IRB) approvals (or an equivalent approval/review based on the requirements of your country or institution) were obtained?
    \item[] Answer: \answerNA{}.
    \item[] Justification: This work does not involve crowdsourcing or research with human subjects, so IRB approval or equivalent review is not applicable to this work.

\item {\bf Declaration of LLM usage}
    \item[] Question: Does the paper describe the usage of LLMs if it is an important, original, or non-standard component of the core methods in this research? Note that if the LLM is used only for writing, editing, or formatting purposes and does \emph{not} impact the core methodology, scientific rigor, or originality of the research, declaration is not required.
    \item[] Answer: \answerYes{}.
    \item[] Justification: LLMs are core components of the proposed routing framework and experimental environment. Their roles as candidate models, high-level router backbone, and Teacher--Student agents are described in Sections~\ref{pre}--\ref{MO} and Appendix~\ref{app:mas_setup}.

\end{enumerate}